\PassOptionsToPackage{nobottomtitles*}{titlesec}
\documentclass[fleqn,10pt]{SelfArx} 

\graphicspath{{figs/}}

\providecommand{\Description}[1]{}

\usepackage[most]{tcolorbox}
\usepackage{fvextra}
\usepackage{pifont}
\usepackage{multirow}
\usepackage{array}
\definecolor{swecream}{RGB}{255,247,236}
\newtcolorbox{observationbox}[1][]{
        colback=envfill,
        colbacktitle=envfill,
        colframe=envborder,
        arc=5pt,
        fontupper=\small,
        fonttitle=\bfseries\color{black},
        boxrule=0.5mm,
        boxsep=1mm,
        width=\linewidth,
        breakable,
        title={Observation \hfill #1},
        rounded corners,
        toptitle=0.7mm,
        bottomtitle=0.7mm
}
\newtcolorbox{goldpatchbox}[1][]{
        colback=goldpatchfill,
        colbacktitle=goldpatchfill,
        colframe=goldpatchborder,
        arc=5pt,
        fontupper=\small,
        fonttitle=\bfseries\color{black},
        boxrule=0.5mm,
        boxsep=1mm,
        width=\linewidth,
        breakable,
        title={Gold Patch \hfill #1},
        rounded corners,
        toptitle=0.7mm,
        bottomtitle=0.7mm
}
\newtcolorbox{issuebox}[1][]{
        colback=issuefill,
        colbacktitle=issuefill,
        colframe=issueborder,
        arc=5pt,
        fontupper=\small,
        fonttitle=\bfseries\color{black},
        boxrule=0.5mm,
        boxsep=1mm,
        width=\linewidth,
        breakable,
        title={Issue \hfill #1},
        rounded corners,
        toptitle=1mm
}
\newtcolorbox{agentbox}[1][]{
        colback=agentfill,
        colbacktitle=agentfill,
        colframe=agentborder,
        arc=5pt,
        fontupper=\small,
        fonttitle=\bfseries\color{black},
        boxrule=0.5mm,
        boxsep=1mm,
        width=\linewidth,
        breakable,
        title={SWE-agent \hfill #1},
        rounded corners,
        toptitle=1mm,
        lower separated=false
}
\newtcolorbox{fileviewerbox}[1]{
        enhanced,
        breakable,
        boxrule = 1.5pt,
        fontupper = \small,
        fonttitle = \bf\color{black},
        arc = 5pt,
        rounded corners,
        colframe = black,
        colbacktitle = swecream,
        colback = swecream,
        title = #1,
        left=4pt 
}
\newtcolorbox{promptbox}[1]{
    enhanced,
    breakable,
    boxrule=1pt,  
    fontupper=\small,
    fonttitle=\bfseries\color{black},
    arc=3pt,  
    rounded corners,
    colframe=black,
    colbacktitle=swecream,
    colback=swecream,
    title=#1,
    left=2mm,  
    right=2mm,  
    top=1mm,  
    bottom=1mm  
}

\definecolor{color1}{RGB}{0,0,90} 
\definecolor{color2}{RGB}{0,20,20} 

\usepackage{hyperref} 
\hypersetup{hidelinks,colorlinks,breaklinks=true,urlcolor=color2,citecolor=color1,linkcolor=color1,bookmarksopen=false,pdftitle={FrameWorkers},pdfauthor={Zhendong Li et al.}}

\Logo{\includegraphics[height=0.8cm]{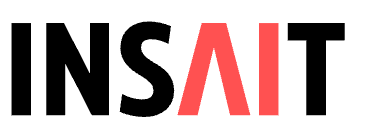}}

\PaperTitle{\textsc{FrameWorkers}: A Dynamic Multi-Agent Framework for AI-Generated Video Production}

\Authors{
Zhendong Li\textsuperscript{1}\quad
Lei Sun\textsuperscript{1}\quad
Letian Shi\textsuperscript{1}\quad
Deheng Zhang\textsuperscript{1}\quad
Ruibo Ming\textsuperscript{1}\quad
Mengshun Hu\textsuperscript{1}\quad
Dannong Xu\textsuperscript{1}\quad
Jian Wang\textsuperscript{2}\quad
Danda Paudel\textsuperscript{1}\quad
Luc Van Gool\textsuperscript{1}\quad
Jinjin Gu\textsuperscript{1}}

\affiliation{\textsuperscript{1}\textit{INSAIT, Sofia University ``St. Kliment Ohridski'', Sofia, Bulgaria}}
\affiliation{\textsuperscript{2}\textit{Snap Inc., New York, USA}}
\affiliation{\textit{Project page}:~~\url{https://frameworkers.insait.ai/}}

\newcommand{\keywordname}{Keywords} 

\Abstract{

Modern video generators excel at synthesizing individual clips, but complete video production requires coordinating a long sequence of interdependent creative steps, including scripting, storyboarding, generation, and editing. It further demands persistent asset management and dynamic task orchestration as intermediate outputs, dependencies, and execution states evolve over time. Existing automated systems typically rely on rigid pipelines that are difficult to adapt to diverse inputs and changing workflows, while general-purpose large language models (LLMs) remain unreliable for long-horizon orchestration and multimodal asset routing. We introduce \textsc{FrameWorkers}, a task-centric and workspace-grounded multi-agent framework for open-ended video production. A central Director formulates video creation as dynamic task management, continuously editing a Task Stack to determine which subtask to execute next and which sub-agent to invoke. An Assistant serves as the execution layer, grounding each selected task in a shared Workspace, retrieving the required assets and context, invoking the assigned sub-agent, and persisting the resulting artifacts. Execution capabilities are exposed through modular sub-agents with registered descriptors, allowing new sub-agents to be integrated without redesigning the orchestration workflow. To improve orchestration reliability, we fine-tune the Director via supervised fine-tuning (SFT) followed by Group Relative Policy Optimization (GRPO) for descriptor-conditioned task routing. Experiments show that \textsc{FrameWorkers} outperforms strong LLM planners in routing accuracy, recovers reliably from runtime failures, generalizes to unseen sub-agents without retraining, and achieves higher end-to-end video quality and broader task coverage than fixed pipelines, single-agent systems, and prior multi-agent approaches.

}

\begin{document}

\setlength{\abovedisplayskip}{10pt}
\setlength{\belowdisplayskip}{10pt}
\setlength{\abovedisplayshortskip}{5pt}
\setlength{\belowdisplayshortskip}{5pt}

\flushbottom

\maketitle


\thispagestyle{empty}


\section{Introduction}
\label{sec:intro}
Recent advances in video generation have substantially expanded the possibilities of AI-assisted content creation.
Powered by large generative models, current systems can synthesize visually rich video clips from text prompts or reference images~\cite{openai2024sora, klingteam2025klingomni, gao2025seedance, wan2025wan}.
These capabilities have significantly lowered the barrier to video creation and made AI-generated content (AIGC) an increasingly promising direction for creative production.
However, in real-world video-making scenarios, generating a single video clip is rarely sufficient.
Short-form videos, advertisements, narrative videos, and short films usually require a complete production process, including script writing, plot development, character design, storyboard planning, dialogue generation, asset organization, video synthesis, editing, and other post-production steps.
Therefore, an important question is how to move beyond isolated clip synthesis toward a complete, automated content production system.

An ideal AIGC video production system should be able to handle diverse user requirements and automatically transform high-level intent into finished video content.
In practice, users may provide only a short textual idea, a detailed script, character portraits, product descriptions, brand constraints, or visual references.
Different inputs naturally require different production workflows.
For instance, when a script is already given, the system should avoid regenerating it; when character images are provided, the system should preserve identity and visual consistency across shots; when the goal is product promotion, the system needs to incorporate product-specific information into the narrative, camera design, and visual style.
Such diversity makes it difficult for a fixed pipeline to cover the wide range of real-world creation tasks.
A dynamic and autonomous multi-agent system is therefore better suited to open-ended and heterogeneous AIGC video production.

Building such a system is challenging.
First, AIGC video production involves many interdependent steps.
Storyboard design depends on the script and narrative structure, character consistency depends on both textual character descriptions and visual assets, and video generation further depends on shot descriptions, prompts, reference images, and style constraints.
Different tasks require different combinations of these steps, and some intermediate steps may fail, requiring replanning, refinement, or additional information.
If a separate workflow is manually designed for each task type, the system quickly suffers from a combinatorial explosion of possible pipelines and becomes difficult to extend.
Second, video production requires extensive asset management.
Text drafts, character profiles, storyboards, reference images, generated clips, editing files, and global context must be correctly stored, retrieved, and passed among sub-agents.
Without a reliable mechanism for asset communication and state tracking, the system can easily lose context, produce inconsistent results, or execute redundant operations.
Third, although pre-trained language models show impressive general planning and reasoning abilities, they are not always reliable when directly used as a director for complex task scheduling.
They may misjudge task priorities, omit necessary steps, or invoke sub-agents at inappropriate moments, which can reduce the stability and efficiency of the entire production process.

We propose \textsc{FrameWorkers}, a dynamic multi-agent framework for automated AIGC video production.
Instead of a fixed pipeline, \textsc{FrameWorkers} formulates video creation as a dynamic task management problem.
The Director agent maintains a Dynamic Task Stack and decides which subtask to perform next and which sub-agent to invoke based on the user request, current task state, available assets, and intermediate execution results.
To support reliable execution, an Assistant agent manages sub-agent interfaces, task inputs and outputs, and a unified asset space containing scripts, character designs, storyboards, reference images, generated clips, and other intermediate results, while preserving cross-step context through Global Memory.
Under this design, existing capabilities can be expanded by adding sub-agent variants. New creative functions can be introduced by registering new sub-agents, without redesigning the workflow. 
We further fine-tune the Director with supervised learning on template-generated, manually verified routing data, followed by Group Relative Policy Optimization (GRPO) on self-generated plans, enabling it to better prioritize tasks, select suitable sub-agents, and construct execution chains under diverse production scenarios.
Experiments show that \textsc{FrameWorkers} supports a broad range of AIGC video production tasks and improves task completion, flexibility, and extensibility over fixed pipelines and direct planning with pretrained models.

\section{Related Work} \label{sec:related}

\noindent\textbf{Video Generation Models and Agents.}\quad
Recent video generation models have achieved remarkable progress in generating high-quality video clips. Lumiere \cite{bar2024lumiere}, VideoPoet \cite{kondratyuk2024videopoet}, CogVideoX \cite{yang2025cogvideox}, and Movie Gen \cite{polyak2024movie} advance space-time generation, multimodal token prediction, diffusion-transformer scaling, and high-resolution multi-task media generation. These backbones improve motion coherence, fidelity, editing, personalization, and audio support, but remain primarily clip-centric and do not handle story decomposition, persistent assets, conditional step skipping, or recovery across a full production workflow.
Long-form systems add explicit narrative structure. VideoStudio \cite{videostudio}, Vlogger \cite{corona2024vlogger}, Animate-A-Story \cite{he2023animate}, and MoPS \cite{ma2024mops} improve multi-scene coherence through scripting, role decomposition, reference conditioning, retrieval-guided motion, or modular story representations. Agentic systems further automate production planning. VideoDirectorGPT \cite{linvideodirectorgpt} expands prompts into structured scene plans, while AesopAgent \cite{wang2024aesopagent}, Mora \cite{yuan2024mora}, StoryAgent \cite{hu2024storyagent}, MM-StoryAgent \cite{xu2025mmstoryagent}, MAViS \cite{wang2026mavis}, AniMaker \cite{shi2025animaker}, and Anim-Director \cite{li2024anim} coordinate specialized roles across scripting, storyboarding, asset and clip generation, and evaluation. These methods improve coherence and automation, but most still assume fixed or task-specific stage orders.
Recent long-form production systems make this limitation clearer. ScriptAgent \cite{mu2026scriptagent}, MovieAgent \cite{wu2025automated}, MovieDreamer \cite{zhao2025moviedreamer}, VideoMemory \cite{zhou2026videomemory}, Camera Artist \cite{hu2026camera}, and CineAGI \cite{xie2026cineagi} introduce executable acting scripts, director-like roles, hierarchical keyframe prediction, dynamic memory, cinematic language, recursive refinement, and audio-visual synchronization. However, they typically optimize a specific formulation or cover only part of asset management and scheduling. Thus, existing work advances long-form generation but leaves dynamic scheduling over optional steps, heterogeneous assets, and failure recovery underexplored.

\noindent\textbf{Agentic Visual Media.}\quad
General visual-agent frameworks show how LLMs can compose tools and expert modules. Visual ChatGPT \cite{wu2023visualchatgpt}, MM-REACT \cite{yang2023mmreact}, ViperGPT \cite{surismenon2023vipergpt}, HuggingGPT \cite{shen2023hugginggpt}, AutoGen \cite{wu2023autogen}, MetaGPT \cite{hong2024metagpt}, and ToolLLM \cite{qin2023toolllm} establish tool invocation, code and model composition, multi-agent communication, role specialization, and API search. They are important precursors to modular media production, but are domain-agnostic and lack a persistent media asset layer spanning scripts, storyboards, references, clips, and editable files.
Media-specific agents begin to add such structure. EditDuet \cite{castaneda2025editduet}, CineAgents \cite{zhang2026cineagent}, UniVA \cite{liang2025univa}, and VQ-Jarvis \cite{zhang2026vqjarvis} use critique loops, hierarchical memory, Plan-and-Act tool use, or operator scheduling for editing, cinematic compilation, general video assistance, and restoration. Agentic designs have also been applied to image-centric media tasks, including autonomous image restoration \cite{chen2024restoreagent, zhu2025agenticir}, photo retouching and composition \cite{chen2026photoartagent, you2026photoframer}, and workflow generation for node-based generation interfaces such as ComfyUI \cite{li2026comfyui}; a recent position paper further argues that agentic systems are a key component of next-generation intelligent image processing \cite{gu2025position}. These systems demonstrate the value of modular tools, shared state, and iterative planning, but remain tied to narrower downstream regimes rather than open-ended AIGC production from diverse specifications to finished video.

Finally, agent-training work motivates fine-tuning a director. AgentTuning \cite{zeng2023agenttuning}, Plan-and-Act \cite{erdogan2025planandact}, and ATLaS \cite{behrouz2025atlas} show that agents benefit from interaction trajectories, planner-executor separation, and learning critical planning steps. For video production, where scheduling errors propagate across story, storyboard, generation, and editing, this suggests a trained director is preferable to relying on prompting alone.

\section{Method}
\label{sec:method}

\begin{figure*}[!t]
    \centering
    \includegraphics[width=\textwidth]{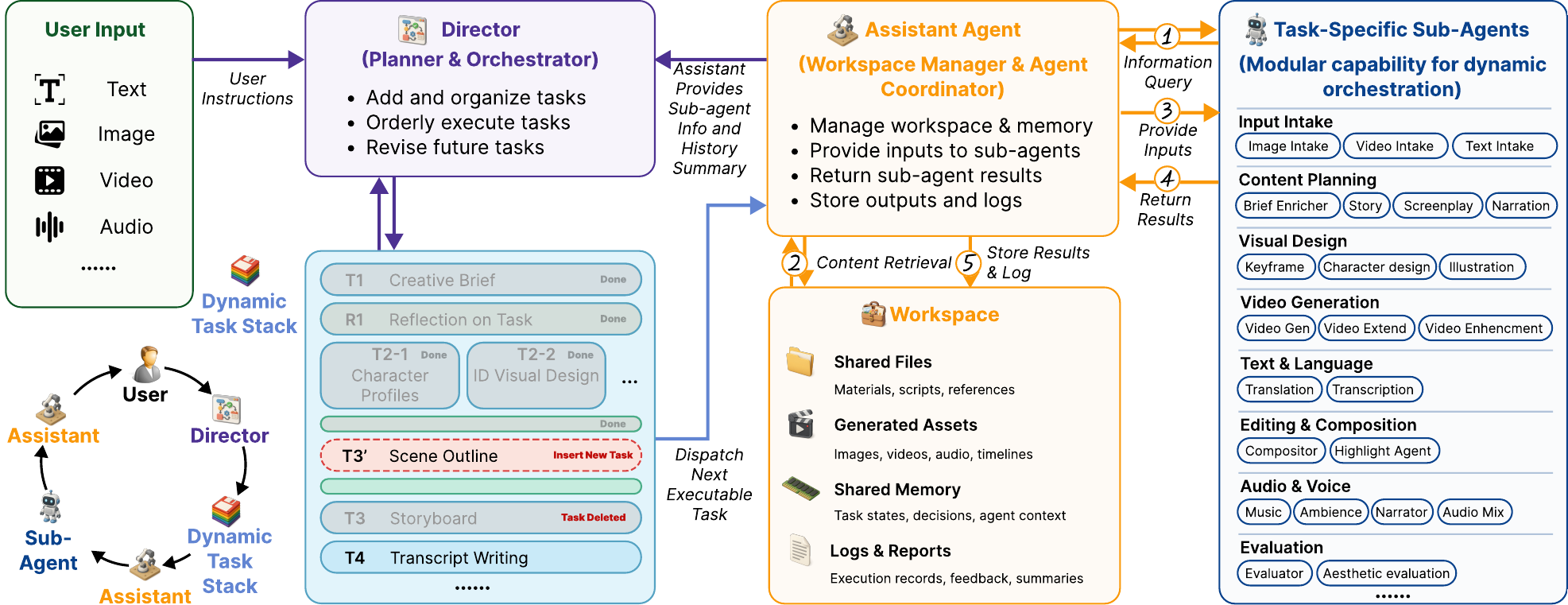}
    \caption{
    Closed-loop orchestration in \textsc{FrameWorkers}. The Director updates the Dynamic Task Stack using the user brief, sub-agent descriptors, and high-level Global Memory. The Assistant executes the next task by resolving inputs from the Workspace, invoking the Sub-agent, checking the result, and writing artifacts and summaries back to the shared state.
    }
        \Description{}
    \label{fig:overview}
\end{figure*}

\subsection{Overview}
\begin{figure*}[p]
    \centering
    \includegraphics[width=\textwidth]{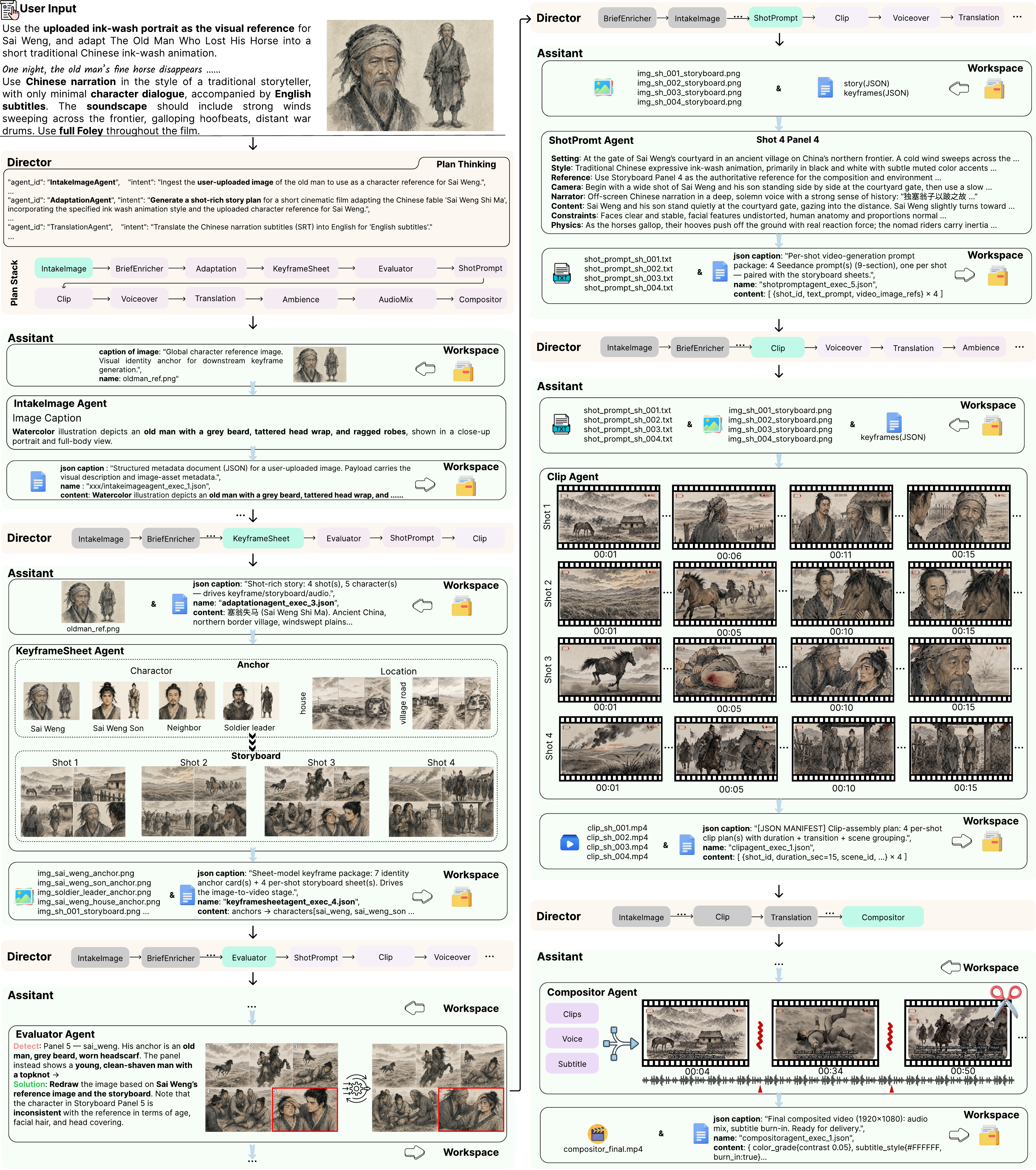}
     \caption{Sample run of \textsc{FrameWorkers} for the user instruction ``The Old Man Who Lost His Horse.'' \textsc{FrameWorkers} takes the user instruction and a reference image of the protagonist as input. The Director constructs an appropriate sub-agent chain to accomplish the task, while the Assistant handles sub-agent execution, asset management, and Global Memory coordination. During execution, if the evaluator detects asset drift—for example, when the appearance of the protagonist in a generated keyframe differs significantly from that in the other keyframes—the corresponding sub-agent is triggered to regenerate the inconsistent output.}
     \Description{}
     \label{fig:example}
\end{figure*}

\textsc{FrameWorkers} is an autonomous multi-agent framework for open-ended video creation.
Unlike systems based on manually specified workflows or fixed pipelines, it closes the loop between planning, execution, evaluation, and replanning.
As shown in \figurename~\ref{fig:overview}, given a user goal and optional multimodal inputs, a central \emph{Director} inspects the user brief, sub-agent descriptors, and the current execution state, then edits a \emph{Dynamic Task Stack} that represents the remaining work.
The next executable task is dispatched to an \emph{Assistant}, which retrieves required artifacts from the shared \emph{Workspace}, invokes the selected self-contained \emph{Sub-agent}, parses and checks the output, and records the result in persistent \emph{Global Memory}.
The execution summary is returned to the Director, which may continue, revise the Task Stack, recover from failures, or incorporate new user instructions.
This separation between strategic planning and concrete execution enables complex multimodal workflows to be assembled dynamically from modular, replaceable sub-agents. A detailed example of our agent system is shown in \figurename~\ref{fig:example}.

\subsection{Multi-Agent Scheme}
\label{sec:subagents}

\textsc{FrameWorkers} organizes execution capabilities as a catalog of independent sub-agents, rather than encoding them in a hand-crafted workflow.
Each sub-agent is a functionally defined capability module, implemented as a simple workflow, an LLM-driven agent, or a more complex agent with iterative self-refinement.
It is exposed through a descriptor specifying its callable identifier, semantic role, required input artifacts, produced output artifacts, and intended trigger conditions.
The Director uses these descriptors to select, order, and revise sub-agent calls, while the Assistant uses them to resolve concrete inputs, invoke the selected sub-agent, parse results, and store artifacts in the shared state.

Because routing depends on semantic descriptions rather than fixed identifiers, new or revised sub-agents can be added without modifying the framework itself.
The right panel of \figurename~\ref{fig:overview} summarizes the sub-agent categories used in this study.
These categories are abstractions, not a one-to-one list of implementations: a category may contain several variants, such as screenplay sub-agents with different prompt designs, narrative preferences, or genre-specific strengths.

Although fixed pipelines could be manually built when the sub-agents and execution order are known, they are brittle under changing user requests, failed intermediate artifacts, or updated sub-agent inventories.
Moreover, connecting heterogeneous sub-agents requires reasoning beyond input-output compatibility, including task intent, artifact availability, and long-range dependencies across video creation.
\textsc{FrameWorkers} therefore uses the Director and Assistant as complementary coordination modules: the Director maintains dynamic task-level scheduling, and the Assistant grounds each selected task in concrete execution.

\subsection{Director and Dynamic Task Stack}
\label{sec:director}
The Director is the high-level orchestration module of \textsc{FrameWorkers}.
It maintains the global task-level plan by editing a \emph{Dynamic Task Stack} according to the user goal, available sub-agent descriptors, current execution state, and new feedback.
Unlike sub-agents, it does not directly execute generation, analysis, or editing models; instead, it decides what should be done next, which sub-agent should handle it, and how the remaining workflow should be revised during execution.

At each planning step, the Director receives a compact planning context. It contains the current user request summary and normalized brief; the full sub-agent descriptor catalog; the Assistant's high-level memory projection, which summarizes completed tasks, generated artifacts, failures, and user-provided files; the current Dynamic Task Stack projection; and any newly appended user instructions.
The descriptor catalog specifies each sub-agent's role, required inputs, expected outputs, and trigger conditions, while the stack projection summarizes planned, pending, active, completed, failed, and cancelled tasks.
Thus, the Director is conditioned not only on the initial request, but also on the evolving state of the session.

The Dynamic Task Stack is an ordered and editable representation of both remaining work and execution history.
Each task entry records a unique identifier, assigned sub-agent, natural-language intent, prerequisites, expected output or state change, execution status, and runtime metadata such as traces or notes.
This representation keeps planning explicit without imposing a fixed pipeline: future tasks can be revised, while completed and failed tasks remain as immutable records of what has already happened.

Rather than producing a static plan once, the Director edits the Task Stack through two primitive operations: \texttt{ADD(task, position)}, which inserts a new task into the stack, and \texttt{DELETE(task\_id)}, which removes a non-executed task.
Task modification is implemented as deletion followed by insertion.
This operation set supports initial planning, task refinement, failure recovery, and adaptation to new user instructions.
The Director may revise the remaining suffix of the stack but cannot erase execution history; this immutability provides recovery semantics and makes the workflow auditable.

After the Task Stack is updated, \textsc{FrameWorkers} selects the next executable task.
A task is executable when its status is \texttt{pending} and its prerequisites are satisfied by completed tasks or available workspace artifacts.
The earliest executable task in the stack is then dispatched to the Assistant.
If no pending task remains and no new user instruction is available, the workflow terminates.
If a task fails, the Assistant writes a failure summary into Global Memory and returns it to the Director.
The Director can then add recovery tasks, delete obsolete pending tasks, or construct a new execution suffix. In our implementation, the fine-tuned Director policy (Section~\ref{sec:director_training}) produces the initial plan, while these runtime revisions are carried out by a general-purpose LLM backend through the same task-stack interface (Section~\ref{subsec:recovery}).

The Dynamic Task Stack gives \textsc{FrameWorkers} substantial flexibility: the system can adapt its plan after intermediate results, failed generations, or changing user intent.
However, this flexibility also makes Director decision-making difficult, since the Director must reason over long-range dependencies, artifact availability, and the functional constraints of heterogeneous sub-agents.
We therefore introduce a dedicated Director training procedure in Section \ref{sec:director_training}.

\subsection{Assistant and Workspace}
\label{sec:assistant}
The Assistant serves as the task execution layer of \textsc{FrameWorkers}. While the Director orchestrates the overall execution process and decides what should be done next, the Assistant determines how each selected task is carried out against the shared state. Given a task from the Dynamic Task Stack, it resolves the required inputs, invokes the assigned sub-agent, validates the returned outputs, persists the artifacts, and returns a compact summary to the Director.

\begin{sloppypar}
The Assistant maintains an abstract \emph{Workspace}, a structured execution environment.
The Workspace contains four persistent components:
\texttt{file\_system} stores textual and structured records such as briefs, scripts, storyboards, annotations, and notes;
\texttt{generated\_assets} stores multimedia artifacts such as images, videos, audio, keyframes, and composited outputs;
\texttt{global\_memory} maintains high-level summaries of completed steps, decisions, produced outputs, current state, failures, uncertainties, user-provided files, and newly appended instructions;
and \texttt{logs} preserve detailed execution traces for debugging, inspection, and recovery.
Workspace records include paths, descriptions, provenance, usage history, generation rationales, and downstream usage information, making intermediate states searchable, reusable, and recoverable.
Rather than exposing all raw files or detailed logs to the Director, the Assistant prepares a compact state projection from \texttt{global\_memory} and relevant Workspace entries, enabling the Director to revise the Task Stack without overloading its context with low-level execution details.
\end{sloppypar}

\begin{figure}[!t]
    \centering
    \includegraphics[width=0.95\linewidth]{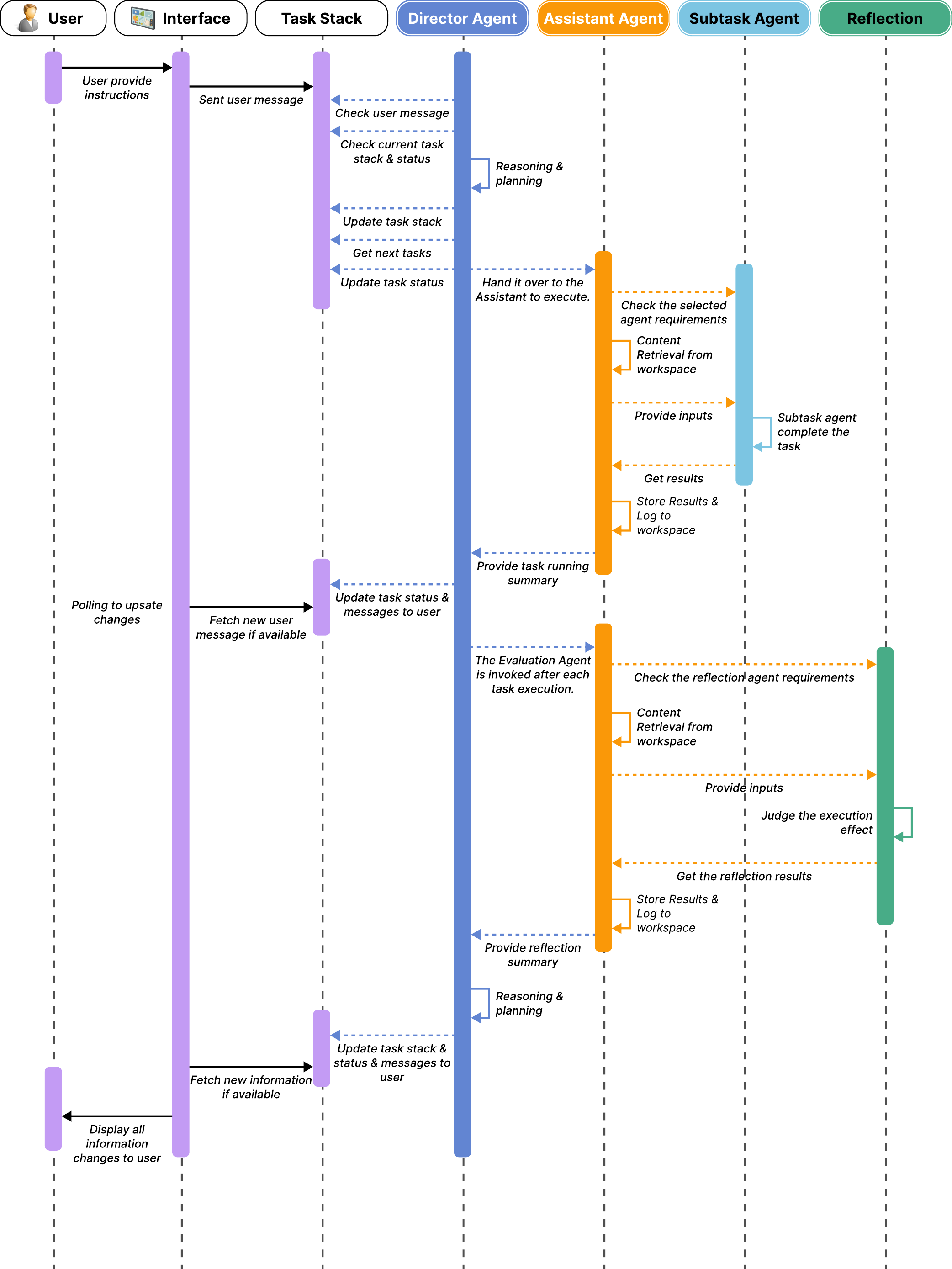}
    \caption{Overall pipeline of the proposed \textsc{FrameWorkers}.}
    \label{fig:flow_chart_supp}
    \Description{}
\end{figure}

As shown in \figurename~\ref{fig:overview} and detailed in \figurename~\ref{fig:flow_chart_supp}, each dispatched task follows a standardized execution procedure.
The Assistant first receives the task intent, execution rationale, assigned sub-agent, expected output, and relevant task-stack context.
The Assistant retrieves the inputs required by the assigned sub-agent from the Workspace—including files, assets, memory summaries, and task metadata—and packages them according to the sub-agent's input schema.
The Assistant subsequently invokes the sub-agent with the resolved input package.
Returned results are parsed into a unified representation and checked for expected fields, file existence, and consistency with the task intent.
Valid outputs are written back as file records or generated assets, high-level events are added to \texttt{global\_memory}, detailed traces are appended to \texttt{logs}, and a compact summary of status, artifacts, state changes, failures, or uncertainties is returned to the Director.

Heterogeneous sub-agents thus remain independently implemented and optimized.
The Assistant acts as an adapter that normalizes their inputs and outputs into the shared Workspace representation under a \emph{strict-out, loose-in} principle.
On the input side, sub-agents consume information through weakly coupled semantic retrieval over Workspace records, assets, and memory summaries.
On the output side, the Assistant enforces structured parsing, file existence checks, metadata recording, and task-level consistency checks before updating shared state.
Thus, sub-agents communicate through high-level artifact semantics rather than tightly synchronized schemas, reducing integration cost and allowing new sub-agents to be added without redesigning the orchestration framework.

\subsection{Director Training}
\label{sec:director_training}

Assistant-side sub-agents solve local tasks with explicit input and output requirements, which pretrained LLMs can often handle with structured parsing and rule-based validation.
The Director instead performs global planning.
Given an underspecified user goal and the full sub-agent descriptor catalog, it must select the necessary sub-agents and order them to satisfy artifact dependencies and video-production logic.
This requires dependency reasoning beyond memorizing frequent sub-agent names or common chain patterns.

We train the Director in two stages.
Supervised fine-tuning (SFT) teaches descriptor-conditioned routing by supervising structured rationales and executable plans.
GRPO then optimizes the properties required for deployment: valid output format, complete sub-agent coverage, and consistent ordering.

\noindent\textbf{Training Task.}\quad
We formulate Director training as descriptor-conditioned routing.
Each instance consists of a prompt $x$ and a target response $y^*$.
The prompt contains the routing rules, the full sub-agent descriptor catalog, and the user goal; the target response contains a routing rationale and an ordered sub-agent plan.
Although the runtime uses a Task Stack, predicting a chain from an empty stack is equivalent to adding a sequence of tasks.
Thus, the key learning problem is to infer the required sub-agents and a valid execution order.

\noindent\textbf{Supervised Fine-Tuning.}\quad
We first train the Director on verified instruction-response pairs
$D_{\mathrm{SFT}}=\{(x_i,y_i^*)\}_{i=1}^{N_{\mathrm{SFT}}}$.
The Director is initialized from Qwen3 models (4B and 8B; 8B unless otherwise stated) and fine-tuned with LoRA adapters.
Each target response $y_i^*=(r_i^*,p_i^*)$ contains a structured routing rationale $r_i^*$ and an executable sub-agent plan $p_i^*$.
The response follows a fixed JSON-style schema:
\begin{quote}
\small
\begin{verbatim}
{"rationale":"...", "plan":[
  {"agent_id":"...", "intent":"..."}]}
\end{verbatim}
\end{quote}
The \texttt{rationale} field decomposes the user request, identifies necessary capabilities, and checks dependency constraints.
The \texttt{plan} field specifies the selected sub-agents and their intended functional roles.
We optimize the standard causal language modeling loss on assistant response tokens only:
\begin{equation}
L_{\mathrm{SFT}}
=
-
\mathbb{E}_{(x,y^*)\sim D_{\mathrm{SFT}}}
\frac{1}{|T(y^*)|}
\sum_{t\in T(y^*)}
\log
\pi_\theta
\left(
y_t^*
\mid
x,y_{<t}^*
\right),
\end{equation}
where $T(y^*)$ denotes the assistant response token positions, $\pi_\theta$ is the Director policy with parameters $\theta$, and prompt tokens are masked out.
This supervision teaches the Director the output format, sub-agent usage semantics, and basic dependencies among multi-step tasks.

\noindent\textbf{GRPO Reward.}\quad
Starting from the SFT policy, we further optimize the Director using GRPO.
For each prompt $x$, the current policy samples $G$ candidate responses
$\{\hat{y}_g\}_{g=1}^{G}$.
Each response is parsed into a rationale and an executable plan $\hat{p}_g$.
A response is valid only if it is parseable, follows the required schema, contains legal sub-agent identifiers, and forms a valid chain.
Invalid responses, including malformed outputs, illegal schemas, and duplicate or unknown sub-agents, receive zero reward.
For valid responses, we score only the executable plan, using sub-agent coverage and dependency-consistent ordering; the rationale is treated as part of the structured output format rather than directly scored for semantic quality. The sequence-level reward for a response $\hat{y}$ is
\begin{equation}
R(\hat{y})
=
R_{\mathrm{val}}(\hat{y})
\left(
\lambda_{\mathrm{val}}
+
\lambda_{\mathrm{cov}} R_{\mathrm{cov}}(\hat{p},p^*)
+
\lambda_{\mathrm{ord}} R_{\mathrm{ord}}(\hat{p},p^*)
\right),
\end{equation}
where $R_{\mathrm{val}}\in\{0,1\}$ indicates response validity, and $\lambda_{\mathrm{val}}$, $\lambda_{\mathrm{cov}}$, and $\lambda_{\mathrm{ord}}$ weight validity, coverage, and ordering.
Let $S(p)$ denote the sub-agents in plan $p$.
The coverage reward is
\begin{equation}
R_{\mathrm{cov}}(\hat{p},p^*)
=
\frac{2\,|S(\hat{p})\cap S(p^*)|}{|S(\hat{p})|+|S(p^*)|}.
\end{equation}
We use the F1 form rather than recall alone. Recall, $|S(\hat{p})\cap S(p^*)|/|S(p^*)|$, does not penalize extra sub-agents: a plan that includes every required sub-agent plus arbitrary others still receives $R_{\mathrm{cov}}=1$, so the policy could raise its reward simply by appending extra sub-agents, since they incur no penalty under recall. F1 penalizes missing and extra sub-agents alike.
Let $E(p^*)$ be the prerequisite edges of the target plan, where $(a,b)\in E(p^*)$ means that $a$ must precede $b$.
The ordering reward is
\begin{equation}
\resizebox{\columnwidth}{!}{$\displaystyle
R_{\mathrm{ord}}(\hat{p},p^*)
=
\begin{cases}
1, & |E(p^*)|=0, \\[1mm]
\frac{1}{|E(p^*)|}
\sum\limits_{(a,b)\in E(p^*)}
\mathbf{1}
\left[
\begin{array}{c}
a,b\in S(\hat{p})\\
\operatorname{pos}_{\hat{p}}(a)<\operatorname{pos}_{\hat{p}}(b)
\end{array}
\right],
& |E(p^*)|>0 .
\end{cases}
$}
\end{equation}
Thus, missing sub-agents receive no ordering credit for the corresponding dependency edges.

\noindent\textbf{GRPO Objective.}\quad
For each prompt group, we normalize rewards among the $G$ sampled responses.
Let $R_g=R(\hat{y}_g)$.
The relative advantage is
\begin{equation}
A_g
=
\frac{R_g-\mathrm{mean}(R_{1:G})}
{\mathrm{std}(R_{1:G})+\epsilon_{\mathrm{adv}}}.
\end{equation}
Here $\epsilon_{\mathrm{adv}}$ is a small constant for numerical stability. This sequence-level advantage is broadcast to all completion tokens.
Because each sampled group is used for a single on-policy gradient step, the importance ratio between the current and sampling policies is identically one, and no clipping is needed.
The objective is a group-normalized policy gradient with a KL anchor to the frozen SFT reference $\pi_{\mathrm{ref}}$:
\begin{equation}
\begin{aligned}
L_{\mathrm{GRPO}}
=
&-\frac{1}{|T|}
\sum_{g}\sum_{t\in T_g}
A_g \log \pi_\theta(\hat{y}_{g,t}\mid x,\hat{y}_{g,<t})
\\
&+
\beta\,
\frac{1}{|T|}
\sum_{g}\sum_{t\in T_g}
\widehat{\mathrm{KL}}_{\mathrm{K3},g,t},
\end{aligned}
\end{equation}
where $T_g$ denotes the completion-token positions of $\hat{y}_g$, $T=\bigcup_g T_g$ ranges over the completion tokens of all sampled responses in the batch, and $\beta$ weights the KL penalty.
We use the non-negative K3 estimator:
%

\begin{equation}
\begin{aligned}
\widehat{\mathrm{KL}}_{\mathrm{K3},g,t} &= \exp(\Delta_{g,t})-\Delta_{g,t}-1, \\
\Delta_{g,t} &= \log \pi_{\mathrm{ref}}(\hat{y}_{g,t}\mid x,\hat{y}_{g,<t}) - \log \pi_\theta(\hat{y}_{g,t}\mid x,\hat{y}_{g,<t}).
\end{aligned}
\end{equation}

When all candidates in a group receive identical rewards, the normalized advantage carries no learning signal, and the group is excluded from the update.
Compared with SFT alone, this direct optimization of executable plan properties improves Chain accuracy and Edit distance on longer chains (Table~\ref{tab:lora_ablation_dual}).

\section{Experiments}
\label{sec:experiments}

\subsection{Data Construction and Task Setup}
\label{sec:exp_datasets}
We evaluate our framework on a structured sub-agent routing task.
Each sample contains a user goal $g$, a target ordered sub-agent chain $\mathbf{c}=(c_1,\ldots,c_n)$, and descriptors of the available sub-agents.
The goals cover diverse AIGC video-production scenarios, including storytelling, highlight generation, multilingual dubbing, style transfer, image-conditioned generation, long-form narration, audio composition, and video extension.
The inventory contains $20$ specialized sub-agents across text, image, video, audio, analysis, and composition.
Each descriptor specifies the agent's required inputs, produced outputs, and trigger conditions, so the Director must construct dependency-consistent chains rather than match surface keywords.
All reference plans are sequential, without branching or parallelism, and each sub-agent appears at most once.

For training, we construct separate SFT and GRPO splits.
The SFT split contains $1{,}404$ instruction-plan pairs across eleven routing categories, with chain lengths typically ranging from $2$ to $10$ agents and long-story goals up to approximately $290$ words.
Ground-truth chains are produced by template-based generation and then manually verified, retaining only fully correct cases.
For each verified chain, we distill a structured chain-of-thought (CoT) reasoning trace from Gemini~2.5 Flash, conditioned on the user goal, descriptor catalog, and target chain.
Thus, each SFT target contains both a rationale and an executable plan, enabling the Director to learn dependency-aware routing rather than only imitate final sub-agent sequences.
The GRPO split contains $1{,}383$ prompts with a chain-shape distribution closely matched to the SFT split, and is used to improve routing consistency and structural correctness under long multi-agent chains.
Regarding data quality and potential bias, the Director does not rely on SFT
alone: SFT bootstraps the structured output format and basic routing behavior,
while GRPO directly optimizes executable properties, including response
validity, sub-agent coverage, and dependency-consistent ordering. The training
data spans eleven routing categories, as shown in Appendix
\figurename~\ref{fig:sft-category}, and the effect of different reasoning
supervision formats is evaluated in
Table~\ref{tab:lora_ablation_dual}. To assess data quality, we manually
inspect a random 20\% sample of the generated instruction-plan pairs,
judging each pair by required capability coverage and dependency-consistent
ordering rather than by exact sequence match, since a production request may
admit multiple valid sub-agent orderings.
Nevertheless, the generated data may still inherit biases from the predefined
routing categories, generation rules, and sub-agent inventory.

We evaluate on two held-out test sets: \textbf{Director-Chain-Easy} (DCE), with $522$ simpler cases, and \textbf{Director-Chain-Hard} (DCH), with $200$ cases containing longer and more structurally demanding chains.
Together, DCE and DCH cover $68$ chain structures, testing both routing accuracy and structural generalization.

\subsection{Director Planning Capability}

We evaluate the Director on DCE and DCH using exact-match chain accuracy (Chain$\uparrow$), per-step routing accuracy (Step$\uparrow$), and set-aware edit distance to the target chain (Edit$\downarrow$). Specifically, let $a=(a_1,\dots,a_n)$ denote the predicted agent plan and $e=(E_1,\dots,E_m)$ the ground-truth chain of \emph{slot sets}, where each $E_j\subseteq\mathcal{A}$ contains all acceptable agents for step $j$. We compute Edit as the standard Levenshtein distance, except that a predicted agent matches a reference slot whenever $a_i\in E_j$:
\[
\begin{aligned}
d_{i,j}
=
\min\!\Bigl\{
&d_{i-1,j}+1,\;
d_{i,j-1}+1,\\
&d_{i-1,j-1}
+\mathbb{1}[a_i\notin E_j]
\Bigr\},
\qquad
d_{i,0}=i,\;
d_{0,j}=j.
\end{aligned}
\]
The final Edit score is $d(a,e)=d_{n,m}$, where lower is better, and $d(a,e)=0$ if and only if the predicted plan has the same length as the reference and every predicted agent belongs to the corresponding reference slot.

We compare with base Qwen3 checkpoints, production-grade non-reasoning chat models, and state-of-the-art reasoning models. All systems receive the same core planning prompt (routing rules, the full descriptor catalog, and the user goal) without few-shot examples or policy hints; our model is the 8B structured-CoT SFT+GRPO variant (SG-SC in Table~\ref{tab:lora_ablation_dual}), which outputs a rationale followed by the plan.


As shown in Table~\ref{tab:director_routing_dual}, SFT+GRPO substantially improves routing over both the original backbones and pretrained reasoning models.
While base Qwen3 checkpoints remain below 30\% Chain accuracy, our model reaches 91.0\% on DCE and 81.5\% on DCH, showing that reliable routing requires task-specific alignment rather than model scale alone.
It also outperforms the strongest pretrained reasoning baseline, Gemini 3 Pro, by $+10.7$ Chain points on DCE and $+34.0$ points on DCH.
The larger DCH margin indicates stronger structural generalization to longer and less frequent chain compositions.
Low Edit scores of 0.15 on DCE and 0.50 on DCH further suggest that the remaining errors stay close to the target chain rather than constituting severe structural failures.
Across model tiers, base models are weak, non-reasoning chat models improve moderately, and reasoning models perform well on DCE but degrade more sharply on DCH.
Our model is the only system above 80\% Chain accuracy on both datasets.

Table~\ref{tab:lora_ablation_dual} analyzes model scale, supervision format, and GRPO refinement.
Scaling from 4B to 8B mainly helps on DCH and on the overall scores (e.g., 68.0 to 81.5 DCH Chain accuracy for SG-SC), whereas DCE is close to saturation for both sizes, indicating that longer and harder chains benefit most from additional planning capacity.
Supervision format is also critical, but its effect depends on the training stage: after SFT alone, direct supervision is the strongest format, whereas after GRPO, structured CoT outperforms both direct supervision and free-form CoT, with the best structured-CoT setting reaching 88.4\% Chain and 93.4\% Step accuracy.
This suggests that, once refined by GRPO, constrained intermediate reasoning stabilizes compositional routing, whereas unconstrained free-form reasoning introduces structural variance.
GRPO improves every SFT checkpoint, with the largest gains under distribution shift, such as $+33.5$ Chain points on DCH for the 8B structured-CoT variant (48.0 to 81.5).
Overall, the results show that the final gains come from structured supervision, reinforcement-based refinement, and sufficient model capacity, and that effective multi-agent routing requires task-specific learning beyond generic LLM reasoning.

\begin{figure*}[!t]
    \centering
    \includegraphics[width=\textwidth]{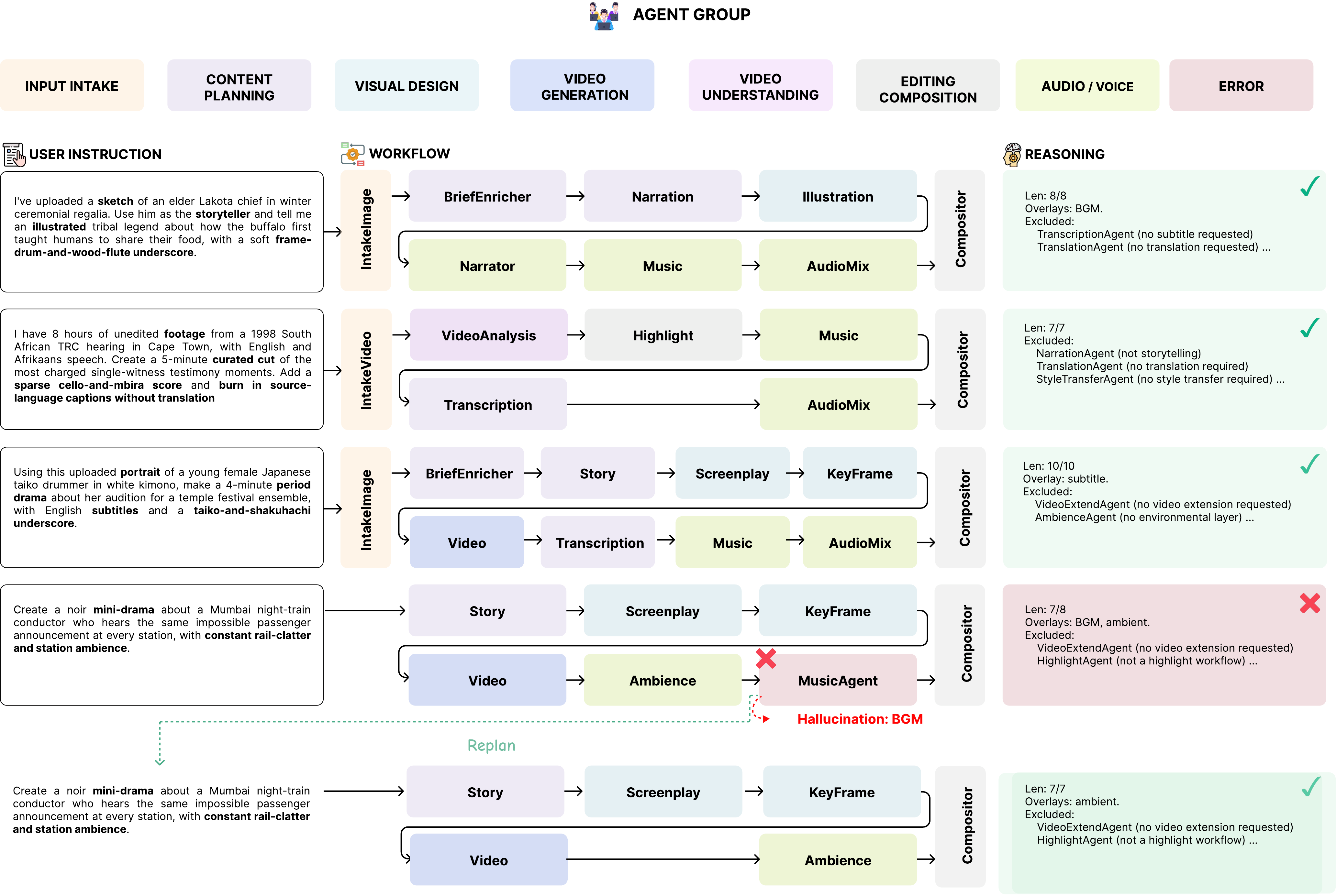}
    \caption[Director Chain]{
    \textbf{Director Chain.}
    Example reasoning traces produced by the Director for multimodal generation prompts.
    Each row shows how an input request is decomposed into a workflow of specialized agents, including input processing, text/planning, visual generation, audio generation, and final composition.
    The reasoning panel records the selected chain length, requested overlays, and excluded agents.
    Correct cases avoid unnecessary modules, while the failure case illustrates an erroneous background-music selection for an ambience-only prompt. After evaluation and replanning, the corrected workflow discards the unnecessary module and retains only the agents required by the request.
}   
\Description{}
    \label{fig:director_chain}
\end{figure*}

Additionally, we provide several example cases in \figurename~\ref{fig:director_chain} to demonstrate the Director's routing capability, including three cases that succeed on the first attempt and one case that initially fails but succeeds after replanning.

\begin{table}[!t]
\centering
\caption{
Director routing accuracy on DCE and DCH.
Ours is Qwen3-8B fine-tuned with structured-CoT SFT and GRPO (SG-SC in Table~\ref{tab:lora_ablation_dual}).
Chain and step accuracy (\%) and Edit distance.}
\label{tab:director_routing_dual}
\small
\setlength{\tabcolsep}{3pt}

\resizebox{\columnwidth}{!}{%
\begin{tabular}{lccc|ccc|ccc}
\toprule
& \multicolumn{3}{c|}{DCE (Easy) (522)} 
& \multicolumn{3}{c|}{DCH (Hard) (200)} 
& \multicolumn{3}{c}{Overall} \\
\cmidrule(lr){2-4} \cmidrule(lr){5-7} \cmidrule(lr){8-10}
\textbf{Model}
& C$\uparrow$ & S$\uparrow$ & E$\downarrow$
& C$\uparrow$ & S$\uparrow$ & E$\downarrow$
& C$\uparrow$ & S$\uparrow$ & E$\downarrow$ \\
\midrule

Qwen3-4B base        & 13.4 & 36.7 & 2.99 & 10.0 & 40.5 & 2.62 & 12.5 & 37.8 & 2.89 \\
Qwen3-8B base        & 29.5 & 52.9 & 1.93 & 11.5 & 42.6 & 1.93 & 24.5 & 50.0 & 1.93 \\
DeepSeek-V3          & 43.9 & 63.2 & 1.11 & 27.5 & 62.3 & 1.28 & 39.4 & 62.9 & 1.16 \\
GPT-4o               & 58.0 & 80.6 & 0.62 & 31.0 & 65.7 & 1.07 & 50.5 & 76.5 & 0.74 \\
Qwen3.6-Plus         & 60.5 & 78.5 & 0.85 & 44.0 & 70.2 & 1.21 & 55.9 & 76.2 & 0.95 \\
Gemini 2.5 Pro       & 72.2 & 86.2 & 0.57 & 51.0 & 78.1 & 0.78 & 66.3 & 83.9 & 0.63 \\
Gemini 3 Pro & 80.3 & 90.6 & 0.32 & 47.5 & 76.4 & 0.84 & 71.2 & 86.7 & 0.46 \\

\midrule
\textbf{Ours}
 & \textbf{91.0} & \textbf{94.4} & \textbf{0.15}
 & \textbf{81.5} & \textbf{90.7} & \textbf{0.50}
 & \textbf{88.4} & \textbf{93.4} & \textbf{0.25} \\

\bottomrule
\end{tabular}%
}
\end{table}

\begin{table}[!t]
\centering
\small
\setlength{\tabcolsep}{4pt}
\caption{
LoRA fine-tuning ablation across DCE and DCH.
SFT denotes supervised fine-tuning and SG denotes SFT+GRPO training.
D / FC / SC denote direct supervision, free-form CoT supervision, and structured-CoT supervision, respectively.
All SG variants are initialized from their corresponding SFT checkpoints.
C / S / E denote Chain accuracy, Step accuracy, and Edit distance, respectively.
Overall is the size-weighted average across both datasets.
}
\label{tab:lora_ablation_dual}

\resizebox{\columnwidth}{!}{%
\begin{tabular}{lccc|ccc|ccc}
\toprule
& \multicolumn{3}{c|}{DCE (Easy) (522)} 
& \multicolumn{3}{c|}{DCH (Hard) (200)} 
& \multicolumn{3}{c}{Overall} \\
\cmidrule(lr){2-4} \cmidrule(lr){5-7} \cmidrule(lr){8-10}

\textbf{Setting}
& C$\uparrow$ & S$\uparrow$ & E$\downarrow$
& C$\uparrow$ & S$\uparrow$ & E$\downarrow$
& C$\uparrow$ & S$\uparrow$ & E$\downarrow$ \\
\midrule

\multicolumn{10}{c}{\textit{4B Models}} \\
\midrule

SFT-D    
& 80.3 & 89.5 & 0.30 
& 66.0 & 85.8 & 0.45 
& 76.2 & 88.3 & 0.34 \\

SFT-FC   
& 66.9 & 83.4 & 0.63 
& 63.0 & 85.4 & 0.52 
& 65.8 & 84.0 & 0.60 \\

SFT-SC   
& 75.1 & 85.8 & 0.46 
& 54.0 & 77.3 & 0.95 
& 69.2 & 83.4 & 0.60 \\

\midrule

SG-D     
& 82.8 & 89.7 & 0.30 
& 67.0 & 87.0 & 0.42 
& 78.4 & 89.0 & 0.33 \\

SG-FC    
& 78.9 & 88.8 & 0.38 
& 65.5 & 85.2 & 0.62 
& 75.1 & 87.7 & 0.45 \\

SG-SC    
& \textbf{92.0} & \textbf{94.7} & 0.16 
& 68.0 & 86.3 & 0.70 
& 85.4 & 92.4 & 0.30 \\

\midrule

\multicolumn{10}{c}{\textit{8B Models}} \\
\midrule

SFT-D    
& 85.2 & 90.4 & 0.34 
& 60.0 & 81.7 & 0.81 
& 78.2 & 87.9 & 0.47 \\

SFT-FC   
& 63.0 & 80.3 & 0.72 
& 62.0 & 80.6 & 0.66 
& 62.7 & 80.4 & 0.70 \\

SFT-SC   
& 78.7 & 85.3 & 0.41 
& 48.0 & 74.8 & 0.94 
& 69.7 & 82.2 & 0.56 \\

\midrule

SG-D     
& 88.7 & 92.4 & 0.23 
& 75.5 & 88.8 & 0.54 
& 85.0 & 91.3 & 0.32 \\

SG-FC    
& 88.5 & 93.1 & 0.22 
& 76.0 & 89.5 & 0.55 
& 85.1 & 92.0 & 0.32 \\

SG-SC
& 91.0 & 94.4 & \textbf{0.15}
& \textbf{81.5} & \textbf{90.7} & \textbf{0.50}
& \textbf{88.4} & \textbf{93.4} & \textbf{0.25} \\

\bottomrule
\end{tabular}%
}
\end{table}

\subsection{Failure Recovery and Replanning}
\label{subsec:recovery}
\textsc{FrameWorkers} operates in a closed loop, where an initially valid plan may
become infeasible after an intermediate failure. We therefore evaluate the
robustness of the complete replanning mechanism under runtime perturbations,
rather than evaluating only initial chain generation from an empty Task
Stack. At each recovery step, the system combines the plan produced by the
fine-tuned Qwen3-8B Director with the structured failure feedback returned by
the Assistant and the functional descriptors of the relevant sub-agents.
Gemini 2.5 Flash is then used as the replanning backend to interpret the failure,
identify the missing or corrective capability, and revise the remaining Task
Stack.

We construct 991 perturbed workflow states using two representative failure
types. For \emph{plan-level perturbations}, we remove a required producer task
while retaining a downstream task that depends on its output. The resulting
Task Stack is structurally incomplete, and successful recovery requires the
replanning mechanism to identify the missing dependency and insert an
appropriate producer before the affected consumer. For
\emph{execution-level perturbations}, a quality gate rejects an artifact
produced by a completed task. The Assistant records the failure type and
returns a structured failure summary, which is combined with the current Task
Stack and the available sub-agent descriptors to determine a
corrective action, such as regeneration or refinement, before downstream
execution continues. In both settings, completed execution history is
preserved, and only the unexecuted portion of the Task Stack can be revised
through the standard \textsc{Add} and \textsc{Delete} operations.

A case counts as recovered when the revised Task Stack
resolves the injected failure, restores all required producer--consumer
dependencies, and yields a valid executable continuation without deleting or
invalidating completed tasks. As reported in Table~\ref{tab:recovery}, the
system recovers 494 of 495 missing-producer cases ($99.8\%$) and 464 of 496
quality-gate failures ($93.5\%$), achieving an overall recovery rate of
$96.7\%$ over 991 cases. The higher success rate for plan-level perturbations
suggests that explicit dependency violations are easier to resolve than
execution-level failures, which require interpreting artifact-quality
feedback and selecting a corrective capability. Together with
the replanning trace in \figurename~\ref{fig:director_chain}, these results show
that the combination of trained initial planning, structured Assistant
feedback, descriptor-grounded capability selection, and LLM-based replanning
provides robust recovery from runtime disruptions.

\begin{table}[!t]
\centering
{
\small
\setlength{\tabcolsep}{4pt}
\caption{Runtime recovery evaluation on 991 perturbed cases. The Director receives no recovery-specific training.}
\label{tab:recovery}
\begin{tabular}{lcc}
\toprule
Error Type & Num & Recovery Success$\uparrow$ \\
\midrule
Missing producer (plan-level)        & 495 & 99.8\% (494) \\
Quality-gate failure (exec-level)    & 496 & 93.5\% (464) \\
\midrule
Overall                              & 991 & 96.7\% (958) \\
\bottomrule
\end{tabular}
}
\end{table}

\subsection{Assistant and Sub-agent Execution Capability}

We evaluate GPT-4o, GPT-4o-mini, Gemini 2.0 Flash, and Gemini 2.5 Flash on a stratified execution suite of 204 cases sampled from DCE and DCH.
We stratify by chain shape and sample up to three cases per structure, yielding a deduplicated benchmark covering 68 distinct chain structures.
All metrics are computed under mock materialization, where external generation APIs are replaced with mock artifacts while preserving the full scheduling and execution logic.
Table~\ref{tab:pass_rework} reports Pass Total, First-shot, Rework, and Latency.
Pass Total measures whether the full chain completes without planning, routing, evaluator, or workspace-level failures; First-shot measures success without rework; Rework measures evaluator-guided recovery; and Latency is the average end-to-end runtime.

As shown in Table~\ref{tab:pass_rework}, execution robustness increases with underlying model capability.
Pass Total rises from 61.76\% for GPT-4o to 81.37\% for GPT-4o-mini, 91.67\% for Gemini 2.0 Flash, and 100\% for Gemini 2.5 Flash.
This suggests that stable long-horizon coordination is primarily a model-capability problem.
Gemini 2.5 Flash completes all 204 cases successfully; 97.55\% succeed on the first attempt, while the remaining 2.45\% are recovered through evaluator-guided rework.
The gap between First-shot and Pass Total shows that evaluator feedback can repair local formatting or structural errors before they propagate, making execution nearly failure-free once the underlying model is sufficiently capable.

Table~\ref{tab:failure_breakdown} decomposes failures into Resolver, Missing, Rework Fail, and Failure Total.
Weaker models mainly fail at the pre-generation planning stage rather than during generation.
GPT-4o and GPT-4o-mini have high Resolver failure rates (32.35\% and 11.76\%), in which downstream agents fail to bind upstream artifacts to required semantic input slots.
Because these errors occur before LLM execution, evaluator-guided rework cannot repair them, making cross-agent semantic grounding the dominant bottleneck.
Their higher Missing failure rates, 5.39\% and 6.86\%, indicate weaker long-range dependency tracking and asset propagation.
Gemini 2.0 Flash reduces Resolver failures to 5.39\%, but still has 2.94\% Rework Fail cases, suggesting that stronger models shift failures from coordination instability toward capability-bound generation failures.
Gemini 2.5 Flash eliminates all observed Resolver, Missing, and Rework Fail cases on this benchmark.
Overall, stable multi-agent execution depends on reliable cross-agent semantic grounding and long-horizon dependency management; once these capabilities are strong enough, evaluator-guided orchestration can sustain highly reliable execution.

\begin{table}[!t]
\centering
\small
\setlength{\tabcolsep}{4pt}
\caption{
Case-level execution robustness and rework recovery analysis across different LLMs.
Pass Total denotes overall successful chain execution,
First-shot denotes success without invoking rework,
Rework denotes recovery through evaluator-guided regeneration,
and Lat. denotes average end-to-end latency.
}
\resizebox{\columnwidth}{!}{%
\begin{tabular}{lcccc}
\toprule
Method
& Pass Total$\uparrow$
& First-shot$\uparrow$
& Rework$\uparrow$ 
& Lat.$\downarrow$ \\
\midrule

GPT-4o-mini 
& 81.37\% (166) 
& 78.43\% (160) 
& 2.94\% (6) 
& 81.1s \\

GPT-4o 
& 61.76\% (126) 
& 55.39\% (113) 
& 6.37\% (13) 
& \textbf{55.1s} \\

Gemini 2.0 Flash 
& 91.67\% (187) 
& 87.25\% (178) 
& 4.41\% (9) 
& 57.9s \\

Gemini 2.5 Flash 
& \textbf{100.0\% (204) }
& \textbf{97.55\% (199) }
& \textbf{2.45\% (5) }
& 137.0s \\

\bottomrule
\end{tabular}%
}

\label{tab:pass_rework}
\end{table}

\begin{table}[!t]
\centering
\small
\setlength{\tabcolsep}{2pt}

\caption{
Failure attribution analysis across different LLMs.
Resolver denotes InputResolver attribution failures,
Missing denotes absent or semantically insufficient upstream artifacts,
and Rework Fail denotes failures unresolved after exhausting the rework loop.
}

\resizebox{\columnwidth}{!}{%
\begin{tabular}{lcccc}
\toprule
Method
& Resolver$\downarrow$
& Missing$\downarrow$ 
& Rework Fail$\downarrow$ 
& Failure Total$\downarrow$ \\
\midrule

GPT-4o-mini
& 11.76\% (24)
& 6.86\% (14)
& 0.00\% (0)
& 18.63\% (38) \\

GPT-4o
& 32.35\% (66)
& 5.39\% (11)
& 0.49\% (1)
& 38.24\% (78) \\

Gemini 2.0 Flash
& 5.39\% (11)
& 0.00\% (0)
& 2.94\% (6)
& 8.33\% (17) \\

Gemini 2.5 Flash
& \textbf{0.00\% (0)}
& \textbf{0.00\% (0)}
& \textbf{0.00\% (0)}
& \textbf{0.00\% (0)} \\

\bottomrule
\end{tabular}%
}

\label{tab:failure_breakdown}
\end{table}

\noindent\textbf{Extensibility to Unseen Sub-agents.}
A key design goal of \textsc{FrameWorkers} is to support plug-and-play
capability expansion without retraining the Director or redesigning existing
workflows. To evaluate this system-level extensibility, we register two
previously unseen sub-agents, ComedyAgent and ScreenwriterAgent, after
training. Each new capability is introduced only through its functional
descriptor and execution wrapper, while the fine-tuned Qwen3-8B Director
remains frozen. The Director must infer the role, required inputs, outputs,
and trigger conditions of each new sub-agent from its descriptor, and
dynamically compose it with existing sub-agents into an executable workflow.
As shown in Table~\ref{tab:newagent}, ComedyAgent and ScreenwriterAgent are
correctly selected in $93.0\%$ and $99.2\%$ of cases, respectively, while the
resulting workflows achieve $99.5\%$ and $88.8\%$ pass rates. These results
demonstrate that \textsc{FrameWorkers} can extend its capability inventory
through descriptor-based registration, with the frozen fine-tuned Director
generalizing to new sub-agents without additional training or changes to the
existing framework. They further suggest that SFT and GRPO
enable the Director to infer a sub-agent's functional role and applicability
from its descriptor, allowing it to generalize beyond the sub-agents seen
during training.

\begin{table}[!t]
\centering
{
\small
\setlength{\tabcolsep}{3pt}
\caption{
System-level extensibility of \textsc{FrameWorkers} to previously unseen
sub-agents. New capabilities are added only through functional descriptors
and execution wrappers, while the SFT- and GRPO-fine-tuned Qwen3-8B Director
remains frozen and must infer their roles from the descriptors. Selection
denotes correct invocation of the new sub-agent; the remaining metrics follow
Table~\ref{tab:pass_rework}.
}
\label{tab:newagent}
\resizebox{\columnwidth}{!}{%
\begin{tabular}{lccccc}
\toprule
New Sub-agent & Selection$\uparrow$ & Pass$\uparrow$ & First-shot$\uparrow$ & Rework$\uparrow$ & Lat.$\downarrow$ \\
\midrule
ComedyAgent (200)        & 93.0\% & 99.5\% & 88.0\% & 11.5\% & 276.3s \\
ScreenwriterAgent (240)  & 99.2\% & 88.8\% & 84.2\% & 4.6\%  & 221.7s \\
\bottomrule
\end{tabular}%
}
}
\end{table}

\subsection{End-to-End Video Evaluation}
\label{subsec:e2e}

\begin{figure*}[!t]
        \centering
     \includegraphics[width=\textwidth]{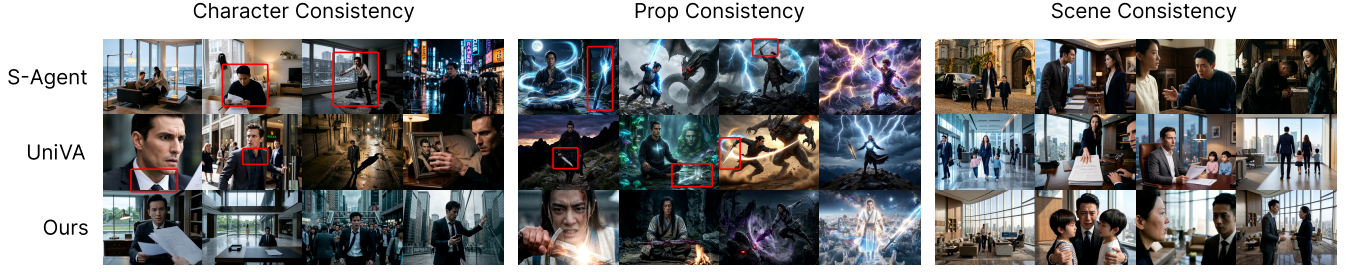}
    \caption{Qualitative comparison demonstrating superior entity consistency. Across all three subclasses (Character, Prop, Scene), \textsc{FrameWorkers} (bottom row) maintains remarkable stability where the baselines fail. 
    The baselines (S-Agent and UniVA) exhibit severe identity drift across different scenarios, including noticeably changing the character’s clothing and overall appearance between scenes (left), morphing the sword into inconsistent shapes and even into entirely different objects across frames (middle), and altering the spatial layout, furniture arrangement, or structural details of the hospital/office environment across shots (right). In contrast, our method preserves the identity and appearance of all entities across distant shots, including character clothing and facial attributes, object geometry, and scene layouts.}
    \Description{}
    \label{fig:vis}
\end{figure*}

\begin{table}[!t]
\centering
\small
\setlength{\tabcolsep}{5pt}
\caption{
Automatic evaluation on the main benchmark.
CLIP denotes the video-text alignment score.
C1--C6 correspond to the MLLM-based evaluation dimensions defined in Section~\ref{subsec:e2e}.
C$_{\text{all}}$ is the average across all six dimensions.
}
\label{tab:video_eval}

\resizebox{\columnwidth}{!}{%
\begin{tabular}{lcccccccc}
\toprule
Method
& CLIP$\uparrow$
& C1$\uparrow$
& C2$\uparrow$
& C3$\uparrow$
& C4$\uparrow$
& C5$\uparrow$
& C6$\uparrow$
& C$_{\text{all}}\uparrow$ \\
\midrule

S-Agent
& 24.94
& 4.50
& 4.55
& 4.50
& 4.60
& 2.70
& 3.65
& 4.08 \\

UniVA
& 24.66
& 4.60
& 4.60
& 4.65
& 4.75
& \textbf{3.25}
& \textbf{3.75}
& 4.27 \\

Ours
& \textbf{25.15}
& \textbf{4.80}
& \textbf{4.90}
& \textbf{4.70}
& \textbf{4.85}
& 3.15
& 3.50
& \textbf{4.32} \\

\bottomrule
\end{tabular}%
}
\end{table}

\providecommand{\cmark}{\ding{51}}
\providecommand{\xmark}{\ding{55}}
\begin{table*}[!t]
\centering
\caption{
Property-level capability comparison of representative academic
video-production systems. \textsc{FrameWorkers} integrates capabilities as
plug-and-play sub-agents and dynamically composes them according to the user
request, available multimodal assets, and current execution state. A checkmark
indicates that the corresponding input configuration or output property is
explicitly supported by an implemented tool or demonstrated workflow; a cross
indicates that such support is not explicitly demonstrated in the
corresponding paper.
}
\label{tab:input_output_comparison}

\setlength{\tabcolsep}{2.7pt}
\renewcommand{\arraystretch}{1.12}

\resizebox{\textwidth}{!}{%
\begin{tabular}{
    @{}
    >{\centering\arraybackslash}m{1.65cm}
    >{\raggedright\arraybackslash}m{9.4cm}
    *{7}{c}
    @{}
}
\toprule

\textbf{Dimension}
&
\makebox[\linewidth][c]{\textbf{Property}}
&
\shortstack[c]{\textbf{FrameWorkers}\\\textbf{(Ours)}}
&
\textbf{UniVA}
&
\shortstack[c]{\textbf{Co-}\\\textbf{Director}}
&
\shortstack[c]{\textbf{Movie}\\\textbf{Agent}}
&
\shortstack[c]{\textbf{Anim-}\\\textbf{Director}}
&
\shortstack[c]{\textbf{Ani}\\\textbf{Maker}}
&
\shortstack[c]{\textbf{MM-Story}\\\textbf{Agent}}
\\

\midrule


\multirow{10}{1.65cm}{
    \centering
    \shortstack[c]{\textbf{Supported}\\\textbf{Inputs}}
}
&
Instruction Only
& \cmark & \cmark & \cmark
& \xmark & \cmark & \cmark & \cmark
\\

&
Instruction + raw Source Text
& \cmark & \cmark & \xmark
& \xmark & \xmark & \xmark & \xmark
\\

&
Instruction + Script
& \cmark & \xmark & \xmark
& \cmark & \xmark & \xmark & \xmark
\\

&
Instruction + Reference Image(s)
& \cmark & \cmark & \cmark
& \cmark & \xmark & \xmark & \xmark
\\

&
Instruction + Video Asset(s)
& \cmark & \cmark & \xmark
& \xmark & \xmark & \xmark & \xmark
\\

&
Instruction + Audio Asset(s)
& \cmark & \cmark & \xmark
& \cmark & \xmark & \xmark & \xmark
\\

&
Instruction + Script + Reference Image(s)
& \cmark & \xmark & \xmark
& \cmark & \xmark & \xmark & \xmark
\\

&
Instruction + Script + Audio Asset(s)
& \cmark & \xmark & \xmark
& \cmark & \xmark & \xmark & \xmark
\\

&
Instruction + Reference Image(s) + Video Asset(s)
& \cmark & \cmark & \xmark
& \xmark & \xmark & \xmark & \xmark
\\

&
Instruction + Script + Reference Image(s) + Audio Asset(s)
& \cmark & \xmark & \xmark
& \cmark & \xmark & \xmark & \xmark
\\

\midrule


\multirow{7}{1.65cm}{
    \centering
    \shortstack[c]{\textbf{Output}\\\textbf{Properties}}
}
&
Multi-shot Dynamic Video
& \cmark & \cmark & \cmark
& \cmark & \cmark & \cmark & \xmark
\\

&
Image-sequence / Storybook Video
& \cmark & \xmark & \xmark
& \xmark & \xmark & \xmark & \cmark
\\

&
Refined Existing Video
& \cmark & \cmark & \xmark
& \xmark & \xmark & \xmark & \xmark
\\

&
Speech / Voice-over Audio
& \cmark & \cmark & \cmark
& \cmark & \cmark & \cmark & \cmark
\\

&
Music / Foley / Sound Effects
& \cmark & \cmark & \cmark
& \xmark & \xmark & \xmark & \cmark
\\

&
Subtitles
& \cmark & \cmark & \xmark
& \cmark & \xmark & \cmark & \xmark
\\

&
Final Audiovisual Composition
& \cmark & \cmark & \cmark
& \cmark & \cmark & \cmark & \cmark
\\

\bottomrule
\end{tabular}%
}

\vspace{2pt}

\begin{minipage}{0.99\textwidth}
\footnotesize
\textit{Notes.}
``Reference Image(s)'' denotes user-provided character, product, logo, style, scene, or keyframe images that serve as visual anchors the generated footage must remain consistent with. ``Video Asset(s)'' denotes user-provided videos consumed either as editable source footage or as motion, camera, temporal, or visual references. ``Audio Asset(s)'' denotes user-provided voice, narration, dialogue, music, or sound-effect recordings consumed as content- or timing-defining input. An input combination is marked as supported only when the listed modalities are jointly consumed within a single, explicitly implemented and demonstrated end-to-end workflow. ``Refined Existing Video'' denotes instruction-guided editing, temporal extension, recomposition, or quality enhancement of a user-supplied video (e.g., highlight recomposition, resolution upscaling, or tail continuation). For grouped audio properties, a checkmark indicates explicit support for at least one listed non-speech audio type (e.g., natively generated foley/ambience, or a muxed background-music track). ``Final Audiovisual Composition'' denotes the automatic assembly and
synchronization of visual clips, voice-over, Foley or music, and subtitles into a single finished deliverable video.
\end{minipage}

\end{table*}

\begin{table}[!t]
\centering
{
\small
\setlength{\tabcolsep}{3.5pt}
\caption{Automatic evaluation on the story-level benchmark (MLLM-as-a-judge). Overall averages C1--C6.}
\label{tab:additional_auto}
\begin{tabular}{lccccccc}
\toprule
Method & Overall$\uparrow$ & C1 & C2 & C3 & C4 & C5 & C6 \\
\midrule
Anim-Director         & 2.88 & 2.95 & 3.35 & 2.70 & 3.50 & 3.55 & 2.55 \\
MovieAgent            & 1.80 & 2.00 & 1.90 & 1.25 & 2.35 & 4.30 & 1.15 \\
\textsc{Ours} & \textbf{4.04} & 4.25 & 4.30 & 3.75 & 4.40 & 4.55 & 3.75 \\
\bottomrule
\end{tabular}
}
\end{table}

\begin{table}[!t]
\centering
{
\small
\setlength{\tabcolsep}{4pt}
\caption{Human evaluation on the story-level benchmark (mean\,$\pm$\,std, 1--5). Twenty independent raters follow the protocol of Section~\ref{subsec:e2e}.}
\label{tab:additional_human}
\resizebox{\columnwidth}{!}{%
\begin{tabular}{lcccc}
\toprule
Method & Instruction$\uparrow$ & Temporal$\uparrow$ & Realism$\uparrow$ & Overall$\uparrow$ \\
\midrule
Anim-Director         & 2.79\,$\pm$\,1.12 & 2.49\,$\pm$\,1.15 & 2.73\,$\pm$\,1.10 & 2.65\,$\pm$\,1.13 \\
MovieAgent            & 2.02\,$\pm$\,0.89 & 2.28\,$\pm$\,1.01 & 2.72\,$\pm$\,0.87 & 2.38\,$\pm$\,0.97 \\
\textsc{Ours} & \textbf{3.50\,$\pm$\,1.05} & \textbf{3.09\,$\pm$\,1.15} & \textbf{3.13\,$\pm$\,1.06} & \textbf{3.21\,$\pm$\,1.10} \\
\bottomrule
\end{tabular}%
}
}
\end{table}

\begin{figure}[!t]
    \centering
    \includegraphics[width=1.05\linewidth]{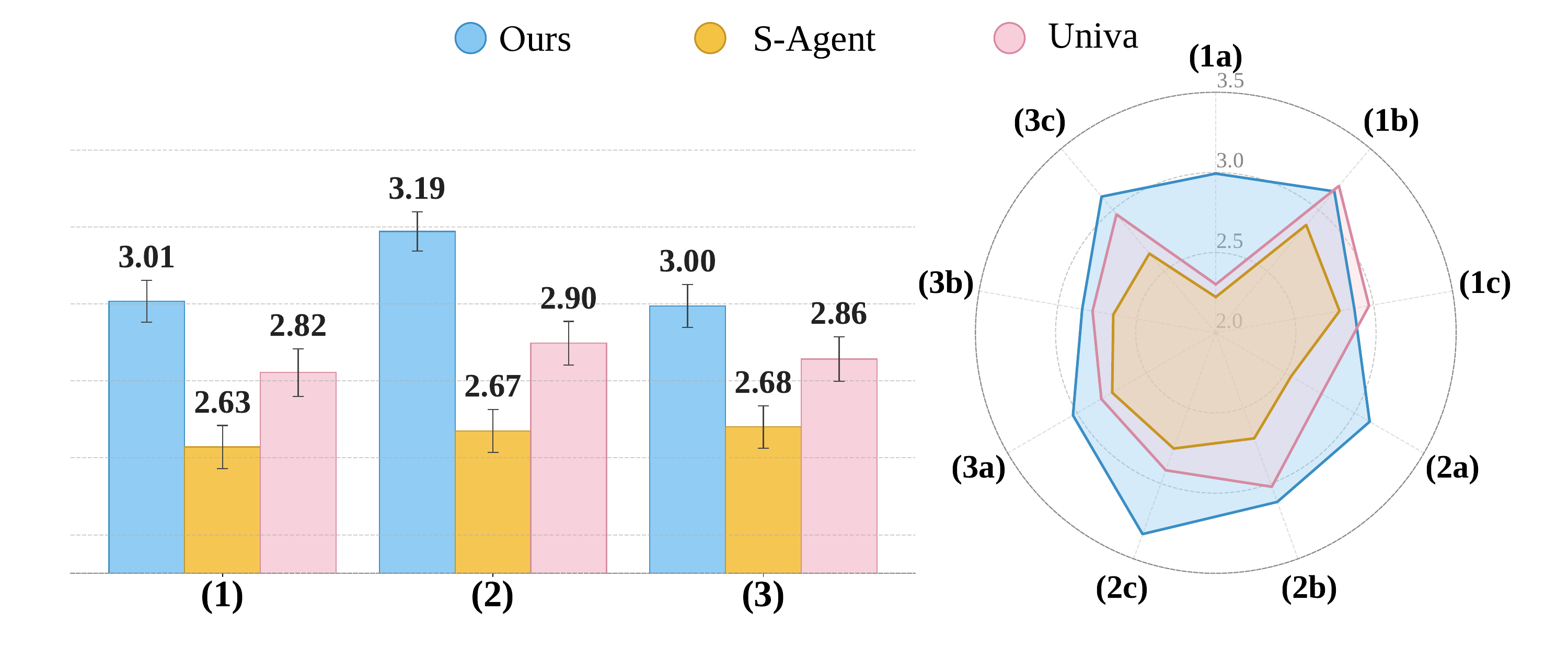}
    \caption{
\textbf{Human evaluation on the main benchmark (against S-Agent and UniVA).}
\textbf{Left:} per-dimension mean $\pm$ SEM, averaged over the three sub-questions in each dimension and across all raters;
\textbf{Right:} per-sub-question mean score for each system.
Our method achieves the highest score on every dimension and on 8 of the 9 sub-questions. The only sub-questions where another system scores higher are (1b) and (1c), where UniVA is slightly higher.
}
\Description{}
    \label{fig:human_eval}
\end{figure}

 We evaluate end-to-end video generation quality under matched prompts by comparing three systems: a single-agent baseline (S-Agent, see Appendix~\ref{sec:details_supp}), UniVA~\cite{liang2025univa}, and \textsc{FrameWorkers}.
All three systems share the same generation backends (\texttt{gemini-3-pro-image-preview} for text-to-image, HunyuanVideo-I2V for image-to-video). Planning in \textsc{FrameWorkers} is performed by our fine-tuned 8B Director (Section~\ref{sec:director_training}); Gemini~2.5 Flash is only the internal backend of the Assistant and sub-agents.
We construct a real-generation benchmark with 20 representative cases, including 10 text-to-video tasks and 10 image-conditioned video-generation tasks.
Each system generates one video per case, yielding 60 videos in total.
This benchmark evaluates the full framework pipeline, including descriptor-driven planning, cross-agent coordination, multimodal artifact grounding, and long-horizon execution consistency.
To broaden the comparison, we further evaluate on a story-level benchmark of 20 cases: five adapted from MovieAgent, five from Anim-Director, and ten newly created cases. We compare against two story-focused baselines, MovieAgent and Anim-Director, and report automatic and human evaluation results below.

\noindent\textbf{Quantitative Evaluation.}
We measure frame-level alignment using the average CLIP ViT-B/32 score between the user instruction and 1-FPS sampled video frames.
We further use Gemini 2.5 Pro as an MLLM-as-a-judge to evaluate the sampled frames and the instruction with a structured prompt (Appendix~\ref{MLLM}).
The judge reports six 1--5 scores: semantic content accuracy (C1), multi-object and spatial relation consistency (C2), action accuracy (C3), story richness and scene diversity (C4), style and cinematic consistency (C5), and overall video-text consistency (C6).

As shown in Table~\ref{tab:video_eval}, \textsc{FrameWorkers} achieves the strongest overall performance across automatic and MLLM-based metrics.
It obtains the highest CLIP score and improves semantic correctness, action consistency, and story richness, reflected by stronger C1--C4 scores.
Although UniVA is slightly better on style consistency and overall video-text consistency, \textsc{FrameWorkers} achieves the best overall average score, with $C_{\text{all}}=4.32$.
These results show that descriptor-driven orchestration and workspace-centered execution improve the semantic fidelity and compositional quality of generated videos. We also include qualitative results in \figurename~\ref{fig:vis}.

\noindent\textbf{Capability Coverage.}
A key design goal of \textsc{FrameWorkers} is to decouple video-production
capabilities from a fixed pipeline and a predefined input-output regime.
Each capability is encapsulated as a self-contained sub-agent and registered
through a semantic descriptor that specifies its functional role, required
inputs, produced outputs, and trigger conditions. This plug-and-play design
allows new capabilities to be incorporated without redesigning the existing
framework. At runtime, the Director dynamically selects and
organizes sub-agents according to the user request, available multimodal
assets, and current execution state, while the Assistant resolves and
transfers heterogeneous artifacts through the shared Workspace. As a result,
different input combinations can induce different production workflows and
deliverable formats within the same system. Table~\ref{tab:input_output_comparison}
shows that this modular and dynamic design gives \textsc{FrameWorkers}
substantially broader input and output coverage than prior systems built
around more fixed production regimes.

On the story-level benchmark, \textsc{FrameWorkers} again achieves the best automatic evaluation performance, outperforming MovieAgent and Anim-Director across all MLLM-judge dimensions (Table~\ref{tab:additional_auto}).

\noindent\textbf{Human Evaluation.}
We conduct a user study on the same 20 cases to assess subjective preference.
To keep each session within approximately 30 minutes, each rater evaluates a deterministic identity-seeded subset of 10 cases.
For each case, the three system outputs are randomly shuffled and shown together with the prompt.
Raters provide 5-point Likert scores for nine sub-questions grouped into three dimensions: instruction following, temporal consistency, and realism.
The full rating instrument is provided in Appendix~\ref{sec:rating-instrument}.
As shown in \figurename~\ref{fig:human_eval}, \textsc{FrameWorkers} outperforms both S-Agent and UniVA across all three dimensions and most sub-questions.
On the story-level benchmark, \textsc{FrameWorkers} likewise achieves the best human evaluation scores across all three dimensions (Table~\ref{tab:additional_human}).

\section{Limitations and Discussion}
\label{sec:limitations}
The main contribution of \textsc{FrameWorkers} is not a new generation
backbone, but a modular coordination mechanism that transforms a fixed
video-production pipeline into a dynamically constructed workflow. Our experiments support this design: the fine-tuned Director improves dependency-aware routing and incorporates newly registered sub-agents without retraining, and task-stack replanning recovers from unseen disruptions. Together, these
results suggest that descriptor-based sub-agent integration and dynamic
task-stack editing form an effective basis for flexible video-production
systems.

At the same time, the current evaluation is conducted within a controlled
system configuration and a predefined sub-agent inventory. Although the
routing, recovery, execution, and extensibility experiments are conducted at a substantially larger scale, the real-generation evaluation remains limited in scale because of the cost of executing complete video-production
pipelines. Therefore, our results demonstrate the effectiveness of our
orchestration design, but do not yet establish robustness across the
full diversity of open-world production requests, sub-agent combinations,
and generation backends. \textsc{FrameWorkers} also inherits the
quality, safety, and reliability limitations of its underlying generative
models.

\section{Conclusion}
\label{sec:conclusion}

In this paper, we presented \textsc{FrameWorkers}, an autonomous multi-agent framework for end-to-end AIGC video production.
By combining Director-based dynamic orchestration with Assistant-supported sub-agent execution and unified asset management, \textsc{FrameWorkers} overcomes the limitations of fixed production pipelines.
Experiments show that it handles diverse user inputs with improved stability, flexibility, and extensibility, providing a scalable foundation for automated AI video creation.
\section{Acknowledgments}
\label{sec:Acknowledge}

This research was partially funded by the Ministry of Education and Science of Bulgaria, through support for INSAIT as part of the Bulgarian National Roadmap for Research Infrastructure. 

\phantomsection


\phantomsection
\bibliographystyle{unsrt}
\bibliography{references}

@article{liang2025univa,
  title={UniVA: Universal Video Agent towards Open-Source Next-Generation Video Generalist},
  author={Liang, Zhengyang and Zhang, Daoan and Zhou, Huichi and Huang, Rui and Li, Bobo and Zhang, Yuechen and Wu, Shengqiong and Wang, Xiaohan and Luo, Jiebo and Liao, Lizi and others},
  journal={arXiv preprint arXiv:2511.08521},
  year={2025}
}

@inproceedings{bar2024lumiere,
  title={Lumiere: A space-time diffusion model for video generation},
  author={Bar-Tal, Omer and Chefer, Hila and Tov, Omer and Herrmann, Charles and Paiss, Roni and Zada, Shiran and Ephrat, Ariel and Hur, Junhwa and Liu, Guanghui and Raj, Amit and others},
  booktitle={SIGGRAPH Asia 2024 Conference Papers},
  pages={1--11},
  year={2024}
}

@inproceedings{kondratyuk2024videopoet,
  title={VideoPoet: a large language model for zero-shot video generation},
  author={Kondratyuk, Dan and Yu, Lijun and Gu, Xiuye and Lezama, Jos{\'e} and Huang, Jonathan and Schindler, Grant and Hornung, Rachel and Birodkar, Vighnesh and Yan, Jimmy and Chiu, Ming-Chang and others},
  booktitle={Proceedings of the 41st International Conference on Machine Learning},
  pages={25105--25124},
  year={2024}
}

@inproceedings{yang2025cogvideox,
  title={Cogvideox: Text-to-video diffusion models with an expert transformer},
  author={Yang, Zhuoyi and Teng, Jiayan and Zheng, Wendi and Ding, Ming and Huang, Shiyu and Xu, Jiazheng and Yang, Yuanming and Hong, Wenyi and Zhang, Xiaohan and Feng, Guanyu and others},
  booktitle={International Conference on Learning Representations},
  volume={2025},
  pages={83048--83077},
  year={2025}
}

@inproceedings{linvideodirectorgpt,
  title={VideoDirectorGPT: Consistent Multi-Scene Video Generation via LLM-Guided Planning},
  author={Lin, Han and Zala, Abhay and Cho, Jaemin and Bansal, Mohit},
  booktitle={First Conference on Language Modeling}
}

@article{polyak2024movie,
  title={Movie gen: A cast of media foundation models},
  author={Polyak, Adam and Zohar, Amit and Brown, Andrew and Tjandra, Andros and Sinha, Animesh and Lee, Ann and Vyas, Apoorv and Shi, Bowen and Ma, Chih-Yao and Chuang, Ching-Yao and others},
  journal={arXiv preprint arXiv:2410.13720},
  year={2024}
}

@article{wang2024aesopagent,
  title={Aesopagent: Agent-driven evolutionary system on story-to-video production},
  author={Wang, Jiuniu and Du, Zehua and Zhao, Yuyuan and Yuan, Bo and Wang, Kexiang and Liang, Jian and Zhao, Yaxi and Lu, Yihen and Li, Gengliang and Gao, Junlong and others},
  journal={arXiv preprint arXiv:2403.07952},
  year={2024}
}

@article{wu2025automated,
  title={Automated movie generation via multi-agent cot planning},
  author={Wu, Weijia and Zhu, Zeyu and Shou, Mike Zheng},
  journal={arXiv preprint arXiv:2503.07314},
  year={2025}
}

@article{hu2024storyagent,
  title={Storyagent: Customized storytelling video generation via multi-agent collaboration},
  author={Hu, Panwen and Jiang, Jin and Chen, Jianqi and Han, Mingfei and Liao, Shengcai and Chang, Xiaojun and Liang, Xiaodan},
  journal={arXiv preprint arXiv:2411.04925},
  year={2024}
}

@inproceedings{zhao2025moviedreamer,
  title={Moviedreamer: Hierarchical generation for coherent long visual sequences},
  author={Zhao, Canyu and Liu, Mingyu and Wang, Wen and Chen, Weihua and Wang, Fan and Chen, Hao and Zhang, Bo and Shen, Chunhua},
  booktitle={International Conference on Learning Representations},
  volume={2025},
  pages={50060--50090},
  year={2025}
}

@article{zhou2026videomemory,
  title={VideoMemory: Toward Consistent Video Generation via Memory Integration},
  author={Zhou, Jinsong and Du, Yihua and Xu, Xinli and Wang, Luozhou and Zhuang, Zijie and Zhang, Yehang and Li, Shuaibo and Hu, Xiaojun and Su, Bolan and Chen, Ying-cong},
  journal={arXiv preprint arXiv:2601.03655},
  year={2026}
}

@article{hu2026camera,
  title={Camera Artist: A Multi-Agent Framework for Cinematic Language Storytelling Video Generation},
  author={Hu, Haobo and Mao, Qi and Li, Yuanhang and Jin, Libiao},
  journal={arXiv preprint arXiv:2604.09195},
  year={2026}
}

@article{xie2026cineagi,
  title={CineAGI: Character-Consistent Movie Creation through LLM-Orchestrated Multi-Modal Generation and Cross-Scene Integration},
  author={Xie, Tianyidan and Huang, Zhentao and Wang, Mingjie and Huang, Xin and Zhou, Jun and Gong, Minglun and Yi, Zili},
  journal={arXiv preprint arXiv:2604.23579},
  year={2026}
}

@article{yuan2024mora,
  title={Mora: Enabling generalist video generation via a multi-agent framework},
  author={Yuan, Zhengqing and Liu, Yixin and Cao, Yihan and Sun, Weixiang and Jia, Haolong and Chen, Ruoxi and Li, Zhaoxu and Lin, Bin and Yuan, Li and He, Lifang and others},
  journal={arXiv preprint arXiv:2403.13248},
  year={2024}
}

@inproceedings{videostudio,
author = {Long, Fuchen and Qiu, Zhaofan and Yao, Ting and Mei, Tao},
title = {VideoStudio: Generating Consistent-Content and Multi-scene Videos},
year = {2024},
isbn = {978-3-031-73026-9},
publisher = {Springer-Verlag},
address = {Berlin, Heidelberg},
url = {https://doi.org/10.1007/978-3-031-73027-6_27},
doi = {10.1007/978-3-031-73027-6_27},
pages = {468–485},
numpages = {18},
location = {Milan, Italy}
}

@inproceedings{corona2024vlogger,
    Author = {Corona, Enric and Zanfir, Andrei and Gabriel Bazavan, Eduard and Kolotouros, Nikos and Alldieck, Thiemo and Sminchisescu, Cristian},
    Title = {VLOGGER: Multimodal Diffusion for Embodied Avatar Synthesis},
    Year = {2024},
    booktitle = {arXiv},
}

@article{he2023animate,
  title={Animate-a-story: Storytelling with retrieval-augmented video generation},
  author={He, Yingqing and Xia, Menghan and Chen, Haoxin and Cun, Xiaodong and Gong, Yuan and Xing, Jinbo and Zhang, Yong and Wang, Xintao and Weng, Chao and Shan, Ying and others},
  journal={arXiv preprint arXiv:2307.06940},
  year={2023}
}

@misc{ma2024mops,
      title={MoPS: Modular Story Premise Synthesis for Open-Ended Automatic Story Generation}, 
      author={Yan Ma and Yu Qiao and Pengfei Liu},
      booktitle={Proceedings of the 62nd Annual Meeting of the Association for Computational Linguistics (Volume 3: System Demonstrations)},
      address={Bangkok, Thailand},
      publisher={Association for Computational Linguistics},
      year={2024},
      url={http://arxiv.org/abs/2406.05690}
}

@misc{xu2025mmstoryagent,
      title={MM-StoryAgent: Immersive Narrated Storybook Video Generation with a Multi-Agent Paradigm across Text, Image and Audio}, 
      author={Xuenan Xu and Jiahao Mei and Chenliang Li and Yuning Wu and Ming Yan and Shaopeng Lai and Ji Zhang and Mengyue Wu},
      year={2025},
      eprint={2503.05242},
      archivePrefix={arXiv},
      primaryClass={cs.CL},
      url={https://arxiv.org/abs/2503.05242}, 
}

@misc{wang2026mavis,
      title={MAViS: A Multi-Agent Framework for Long-Sequence Video Storytelling}, 
      author={Qian Wang and Ziqi Huang and Ruoxi Jia and Paul Debevec and Ning Yu},
      year={2026},
      eprint={2508.08487},
      archivePrefix={arXiv},
      primaryClass={cs.CV},
      url={https://arxiv.org/abs/2508.08487}, 
}

@inproceedings{li2024anim,
  title={Anim-director: A large multimodal model powered agent for controllable animation video generation},
  author={Li, Yunxin and Shi, Haoyuan and Hu, Baotian and Wang, Longyue and Zhu, Jiashun and Xu, Jinyi and Zhao, Zhen and Zhang, Min},
  booktitle={SIGGRAPH Asia 2024 Conference Papers},
  pages={1--11},
  year={2024}
}

@inproceedings{shi2025animaker,
  title={AniMaker: Multi-Agent Animated Storytelling with MCTS-Driven Clip Generation},
  author={Shi, Haoyuan and Li, Yunxin and Chen, Xinyu and Wang, Longyue and Hu, Baotian and Zhang, Min},
  booktitle={Proceedings of the SIGGRAPH Asia 2025 Conference Papers},
  pages={1--11},
  year={2025}
}

@misc{mu2026scriptagent,
      title={The Script is All You Need: An Agentic Framework for Long-Horizon Dialogue-to-Cinematic Video Generation}, 
      author={Chenyu Mu and Xin He and Qu Yang and Wanshun Chen and Jiadi Yao and Huang Liu and Zihao Yi and Bo Zhao and Xingyu Chen and Ruotian Ma and Fanghua Ye and Erkun Yang and Cheng Deng and Zhaopeng Tu and Xiaolong Li and Linus},
      year={2026},
      eprint={2601.17737},
      archivePrefix={arXiv},
      primaryClass={cs.CV},
      url={https://arxiv.org/abs/2601.17737}, 
}

@article{yang2023mmreact,
  author      = {Zhengyuan Yang* and Linjie Li* and Jianfeng Wang* and Kevin Lin* and Ehsan Azarnasab* and Faisal Ahmed* and Zicheng Liu and Ce Liu and Michael Zeng and Lijuan Wang\^},
  title       = {MM-REACT: Prompting ChatGPT for Multimodal Reasoning and Action},
  publisher   = {arXiv},
  year        = {2023},
}

@article{surismenon2023vipergpt,
    title={ViperGPT: Visual Inference via Python Execution for Reasoning},
    author={D\'idac Sur\'is and Sachit Menon and Carl Vondrick},
    journal={Proceedings of IEEE International Conference on Computer Vision (ICCV)},
    year={2023}
}

@inproceedings{shen2023hugginggpt,
  author = {Shen, Yongliang and Song, Kaitao and Tan, Xu and Li, Dongsheng and Lu, Weiming and Zhuang, Yueting},
  booktitle = {Advances in Neural Information Processing Systems},
  title = {HuggingGPT: Solving AI Tasks with ChatGPT and its Friends in HuggingFace},
  year = {2023}
}

@misc{wu2023autogen,
      title={AutoGen: Enabling Next-Gen LLM Applications via Multi-Agent Conversation}, 
      author={Qingyun Wu and Gagan Bansal and Jieyu Zhang and Yiran Wu and Beibin Li and Erkang Zhu and Li Jiang and Xiaoyun Zhang and Shaokun Zhang and Jiale Liu and Ahmed Hassan Awadallah and Ryen W White and Doug Burger and Chi Wang},
      year={2023},
      eprint={2308.08155},
      archivePrefix={arXiv},
      primaryClass={cs.AI},
      url={https://arxiv.org/abs/2308.08155}, 
}

@inproceedings{hong2024metagpt,
      title={Meta{GPT}: Meta Programming for A Multi-Agent Collaborative Framework},
      author={Sirui Hong and Mingchen Zhuge and Jonathan Chen and Xiawu Zheng and Yuheng Cheng and Jinlin Wang and Ceyao Zhang and Zili Wang and Steven Ka Shing Yau and Zijuan Lin and Liyang Zhou and Chenyu Ran and Lingfeng Xiao and Chenglin Wu and J{\"u}rgen Schmidhuber},
      booktitle={The Twelfth International Conference on Learning Representations},
      year={2024},
      url={https://openreview.net/forum?id=VtmBAGCN7o}
}

@misc{qin2023toolllm,
      title={ToolLLM: Facilitating Large Language Models to Master 16000+ Real-world APIs}, 
      author={Yujia Qin and Shihao Liang and Yining Ye and Kunlun Zhu and Lan Yan and Yaxi Lu and Yankai Lin and Xin Cong and Xiangru Tang and Bill Qian and Sihan Zhao and Runchu Tian and Ruobing Xie and Jie Zhou and Mark Gerstein and Dahai Li and Zhiyuan Liu and Maosong Sun},
      year={2023},
      eprint={2307.16789},
      archivePrefix={arXiv},
      primaryClass={cs.AI}
}

@misc{castaneda2025editduet,
      title={EditDuet: A Multi-Agent System for Video Non-Linear Editing}, 
      author={Marcelo Sandoval-Castaneda and Bryan Russell and Josef Sivic and Gregory Shakhnarovich and Fabian Caba Heilbron},
      year={2025},
      eprint={2509.10761},
      archivePrefix={arXiv},
      primaryClass={cs.CV},
      url={https://arxiv.org/abs/2509.10761}, 
}

@misc{zhang2026cineagent,
      title={A Benchmark and Multi-Agent System for Instruction-driven Cinematic Video Compilation}, 
      author={Peixuan Zhang and Chang Zhou and Ziyuan Zhang and Hualuo Liu and Chunjie Zhang and Jingqi Liu and Xiaohui Zhou and Xi Chen and Shuchen Weng and Si Li and Boxin Shi},
      year={2026},
      eprint={2604.10456},
      archivePrefix={arXiv},
      primaryClass={cs.CV},
      url={https://arxiv.org/abs/2604.10456}, 
}

@misc{zhang2026vqjarvis,
      title={VQ-Jarvis: Retrieval-Augmented Video Restoration Agent with Sharp Vision and Fast Thought}, 
      author={Xuanyu Zhang and Weiqi Li and Qunliang Xing and Jingfen Xie and Bin Chen and Junlin Li and Li Zhang and Jian Zhang and Shijie Zhao},
      year={2026},
      eprint={2603.22998},
      archivePrefix={arXiv},
      primaryClass={cs.CV},
      url={https://arxiv.org/abs/2603.22998}, 
}

@misc{zeng2023agenttuning,
      title={AgentTuning: Enabling Generalized Agent Abilities for LLMs},
      author={Aohan Zeng and Mingdao Liu and Rui Lu and Bowen Wang and Xiao Liu and Yuxiao Dong and Jie Tang},
      year={2023},
      eprint={2310.12823},
      archivePrefix={arXiv},
      primaryClass={cs.CL}
}

@inproceedings{
erdogan2025planandact,
title={Plan-and-Act: Improving Planning of Agents for Long-Horizon Tasks},
author={Lutfi Eren Erdogan and Hiroki Furuta and Sehoon Kim and Nicholas Lee and Suhong Moon and Gopala Anumanchipalli and Kurt Keutzer and Amir Gholami},
booktitle={Forty-second International Conference on Machine Learning},
year={2025},
url={https://openreview.net/forum?id=ybA4EcMmUZ}
}

@misc{behrouz2025atlas,
      title={ATLAS: Learning to Optimally Memorize the Context at Test Time}, 
      author={Ali Behrouz and Zeman Li and Praneeth Kacham and Majid Daliri and Yuan Deng and Peilin Zhong and Meisam Razaviyayn and Vahab Mirrokni},
      year={2025},
      eprint={2505.23735},
      archivePrefix={arXiv},
      primaryClass={cs.CL},
      url={https://arxiv.org/abs/2505.23735}, 
}

@misc{openai2024sora,
  title        = {Video Generation Models as World Simulators},
  author       = {OpenAI},
  year         = {2024},
  month        = feb,
  howpublished = {\url{https://openai.com/index/video-generation-models-as-world-simulators/}},
  note         = {Technical report introducing Sora}
}

@misc{klingteam2025klingomni,
      title={Kling-Omni Technical Report}, 
      author={Kling Team},
      year={2025},
      eprint={2512.16776},
      archivePrefix={arXiv},
      primaryClass={cs.CV},
      url={https://arxiv.org/abs/2512.16776}, 
}

@misc{gao2025seedance,
      title={Seedance 1.0: Exploring the Boundaries of Video Generation Models}, 
      author={Yu Gao and Haoyuan Guo and Tuyen Hoang and Weilin Huang and Lu Jiang and Fangyuan Kong and Huixia Li and Jiashi Li and Liang Li and Xiaojie Li and Xunsong Li and Yifu Li and Shanchuan Lin and Zhijie Lin and Jiawei Liu and Shu Liu and Xiaonan Nie and Zhiwu Qing and Yuxi Ren and Li Sun and Zhi Tian and Rui Wang and Sen Wang and Guoqiang Wei and Guohong Wu and Jie Wu and Ruiqi Xia and Fei Xiao and Xuefeng Xiao and Jiangqiao Yan and Ceyuan Yang and Jianchao Yang and Runkai Yang and Tao Yang and Yihang Yang and Zilyu Ye and Xuejiao Zeng and Yan Zeng and Heng Zhang and Yang Zhao and Xiaozheng Zheng and Peihao Zhu and Jiaxin Zou and Feilong Zuo},
      year={2025},
      eprint={2506.09113},
      archivePrefix={arXiv},
      primaryClass={cs.CV},
      url={https://arxiv.org/abs/2506.09113}, 
}

@misc{wan2025wan,
      title={Wan: Open and Advanced Large-Scale Video Generative Models}, 
      author={Team Wan},
      year={2025},
      eprint={2503.20314},
      archivePrefix={arXiv},
      primaryClass={cs.CV},
      url={https://arxiv.org/abs/2503.20314}, 
}

@misc{wu2023visualchatgpt,
      title={Visual ChatGPT: Talking, Drawing and Editing with Visual Foundation Models},
      author={Chenfei Wu and Shengming Yin and Weizhen Qi and Xiaodong Wang and Zecheng Tang and Nan Duan},
      year={2023},
      eprint={2303.04671},
      archivePrefix={arXiv},
      primaryClass={cs.CV}
}

@inproceedings{chen2024restoreagent,
  title={RestoreAgent: Autonomous Image Restoration Agent via Multimodal Large Language Models},
  author={Chen, Haoyu and Li, Wenbo and Gu, Jinjin and Ren, Jingjing and Chen, Sixiang and Ye, Tian and Pei, Renjing and Zhou, Kaiwen and Song, Fenglong and Zhu, Lei},
  booktitle={Advances in Neural Information Processing Systems},
  year={2024}
}

@inproceedings{zhu2025agenticir,
  title={An Intelligent Agentic System for Complex Image Restoration Problems},
  author={Zhu, Kaiwen and Gu, Jinjin and You, Zhiyuan and Qiao, Yu and Dong, Chao},
  booktitle={International Conference on Learning Representations},
  year={2025}
}

@inproceedings{chen2026photoartagent,
  title={PhotoArtAgent: Intelligent Photo Retouching with Language Model-Based Artist Agents},
  author={Chen, Haoyu and Tao, Keda and Wang, Yizao and Wang, Xinlei and Zhu, Lei and Gu, Jinjin},
  booktitle={Findings of the IEEE/CVF Conference on Computer Vision and Pattern Recognition},
  year={2026}
}

@inproceedings{you2026photoframer,
  title={PhotoFramer: Multi-modal Image Composition Instruction},
  author={You, Zhiyuan and Wang, Ke and Zhang, He and Cai, Xin and Gu, Jinjin and Xue, Tianfan and Dong, Chao and Zhang, Zhoutong},
  booktitle={Proceedings of the IEEE/CVF Conference on Computer Vision and Pattern Recognition},
  year={2026}
}

@inproceedings{li2026comfyui,
  title={Knowledge-Centric Agents for Workflow Generation in ComfyUI},
  author={Li, Zhendong and Sun, Lei and Ming, Ruibo and Zhang, He and Paudel, Danda Pani and Van Gool, Luc and Gu, Jinjin},
  booktitle={European Conference on Computer Vision},
  year={2026}
}

@article{gu2025position,
  title={Position: Agentic Systems Constitute a Key Component of Next-Generation Intelligent Image Processing},
  author={Gu, Jinjin},
  journal={arXiv preprint arXiv:2505.16007},
  year={2025}
}

\appendix
\cleardoublepage

\section*{Appendix}

This appendix provides extended implementation details, expanded experimental analysis, and a full description of our evaluation protocols to complement the main paper.

In Section~\ref{sec:workflow_supp}, we provide a detailed procedural walkthrough of the \textsc{FrameWorkers} pipeline, illustrating the closed-loop orchestration between the Director, Assistant, and Workspace. Section~\ref{sec:robustness} presents an in-depth robustness analysis, offering a per-segment breakdown of failure modes across the four evaluated backends, with emphasis on the two Gemini versions. For reproducibility, Section~\ref{sec:details_supp} elaborates on implementation details, including the design of our baseline (S-Agent) and the complete prompt template used for our MLLM-as-a-judge evaluation. Finally, Section~\ref{sec:user-study} details our user study methodology, providing the full rating instrument and operational rubrics used by human evaluators.

\section{More Details About \textsc{FrameWorkers} Pipeline}
\label{sec:workflow_supp}

As shown in \figurename~\ref{fig:flow_chart_supp}, \textsc{FrameWorkers} operates as a closed-loop, autonomous multi-agent framework for open-ended video creation, continuously cycling through planning, execution, evaluation, and replanning. The workflow begins when a user provides instructions through the system interface, which are then forwarded to the central Director. The Director evaluates this new user message alongside the current status of the Task Stack and the overall execution state. Utilizing a catalog of available sub-agents, the Director performs high-level reasoning and planning to determine the sequence of necessary operations, translating this plan into an actionable format by dynamically updating the Task Stack.

Once the Task Stack is updated, the Director identifies the next executable task and dispatches it to the Assistant. Serving as the execution runtime, the Assistant inspects the input requirements of the sub-agent assigned to the job and retrieves the necessary content and artifacts from the shared Workspace. It then provides these formatted inputs to the sub-agent, which executes its generation or processing task. After the sub-agent completes its operation, the Assistant retrieves the results, stores them, and logs the detailed execution traces back into the Workspace.

Following this concrete execution, an evaluation phase is triggered. The Assistant gathers the necessary context from the Workspace and applies the evaluator's checks to the produced artifacts (schema and structural validation, content-level quality checks, and asset existence checks) and stores the evaluation results in the shared Workspace as well. Finally, the Assistant compiles a compact task and evaluation summary and returns it to the Director. Armed with this updated state and feedback, the Director engages in further reasoning to decide if the plan needs revision, failure recovery, or continuation, ultimately updating the Task Stack and preparing status messages. The interface periodically polls for these updates, fetches new user information if available, and displays all generated changes to the user, continuing this closed-loop cycle until the user's overarching goal is achieved.

\section{Robustness and Failure Modes}
\label{sec:robustness}

Table~\ref{tab:segment_breakdown} decomposes failure distributions across the four backends of Section~\ref{sec:experiments}; we focus on the two Gemini variants. For Gemini 2.0 Flash, failures are spread over several categories, most notably \textit{audio} (66.7\% pass), \textit{style} (85.7\%), \textit{storytelling} (86.1\%), \textit{bilingual} (88.9\%), \textit{cr} (91.7\%), and \textit{complex} (92.9\%), suggesting that long-horizon multimodal coordination remains challenging under deeper generation chains. Step-level failures are likewise distributed across multiple agents, particularly BriefEnricherAgent (88.1\%), AudioMixAgent (93.8\%), and MusicAgent (96.9\%), indicating instability in long-context planning and audio-scene synchronization.

In contrast, Gemini 2.5 Flash passes all 204 cases and completes every planned step: the five cases that required rework (Table~\ref{tab:pass_rework}) are all recovered within the rework loop, so no category or agent falls below 100\%. This indicates that Gemini 2.5 Flash largely resolves the cross-agent consistency and long-horizon coordination issues observed in Gemini 2.0 Flash.

\begin{table}[!ht]
\centering
\small
\setlength{\tabcolsep}{2pt}
\caption{
Per-segment breakdown (worst-first ordering).
We report case-level pass rate and step-level completion rate (completed over attempted steps); $N$ is the number of cases or of planned steps. Segments and agents with perfect accuracy across all backends are grouped as \textit{Other}.
}
\label{tab:segment_breakdown}
\resizebox{0.45\textwidth}{!}{
\begin{tabular}{llccccc}
\toprule
\textbf{Segment} 
& \textbf{Metric} 
& \textbf{N} 
& \textbf{GPT-4o} 
& \textbf{GPT-4o-mini} 
& \textbf{Gemini-2.0} 
& \textbf{Gemini-2.5} \\
\midrule
highlight & case & 24 & 0.00 & 91.67 & 95.83 & -- \\
cr & case & 12 & 41.67 & 75.00 & 91.67 & -- \\
bilingual & case & 9 & 55.56 & 88.89 & 88.89 & -- \\
complex & case & 42 & 61.90 & 88.10 & 92.86 & -- \\
extend & case & 21 & 66.67 & 85.71 & -- & -- \\
sub & case & 6 & 66.67 & 66.67 & -- & -- \\
sub\_vid & case & 6 & 66.67 & -- & -- & -- \\
storytelling & case & 36 & 75.00 & 55.56 & 86.11 & -- \\
intake\_img & case & 21 & 80.95 & 80.95 & 95.24 & -- \\
style & case & 21 & 85.71 & 90.48 & 85.71 & -- \\
audio & case & 6 & -- & -- & 66.67 & -- \\
\midrule
HighlightAgent & step & 33 & 0.00 & 96.97 & -- & -- \\
TranslationAgent & step & 54 & 74.19 & -- & -- & -- \\
NarrationAgent & step & 36 & 80.00 & -- & -- & -- \\
StoryAgent & step & 66 & 83.87 & -- & -- & -- \\
BriefEnricherAgent & step & 42 & 88.10 & -- & 88.10 & -- \\
VideoExtendAgent & step & 45 & 93.02 & -- & -- & -- \\
KeyFrameAgent & step & 66 & 94.00 & -- & -- & -- \\
MusicAgent & step & 102 & 95.59 & -- & 96.94 & -- \\
VideoAgent & step & 66 & 95.74 & -- & 98.46 & -- \\
ScreenplayAgent & step & 66 & 96.15 & -- & -- & -- \\
AudioMixAgent & step & 120 & 98.75 & 76.27 & 93.75 & -- \\
CompositorAgent & step & 192 & 99.15 & 95.65 & -- & -- \\
AmbienceAgent & step & 63 & -- & 96.83 & 98.33 & -- \\
Other (step) & step & -- & 100.0 & 100.0 & 100.0 & 100.0 \\
\bottomrule
\end{tabular}
}
\end{table}

\section{More Implementation Details}
\label{sec:details_supp}




To establish a solid baseline for comparison, we implement a single-agent system, denoted as {S-Agent}. Unlike our proposed framework, the S-Agent operates on a straightforward, sequential pipeline without multi-agent collaboration. Given a user instruction, the S-Agent utilizes a single query to an MLLM to handle the planning phase. Specifically, the MLLM is prompted to decompose the high-level instruction into a sequential list of individual shots and simultaneously generate detailed textual prompts for the keyframe of each shot. Following this planning stage, the system executes a two-step visual generation process. First, a Text-to-Image (T2I) model is employed to synthesize the keyframes based on the generated shot prompts. Subsequently, an Image-to-Video (I2V) model takes these synthesized keyframes as starting points to render the final video segments. This I2V process guarantees visual quality and spatio-temporal consistency strictly within the boundaries of each individual shot.

\subsection{Sub-agent Descriptors}
\label{sec:descriptors_supp}

Each sub-agent is exposed to the Director through a plain-text descriptor containing three fields.
\texttt{Inputs} specifies the artifact labels consumed by the sub-agent.
Each input entry includes its cardinality (\texttt{single} or \texttt{collection}), whether it is \texttt{required} or \texttt{optional}, and a natural-language description that specifies the expected artifact and, when needed, provides guidance for resolving it from the Workspace.
The Assistant uses this information to retrieve and select concrete inputs from the Workspace.
\texttt{Output} specifies the artifact produced by the sub-agent, while \texttt{Purpose / Trigger} describes when the sub-agent should be invoked.

Importantly, each descriptor describes only its own sub-agent and does not explicitly refer to other sub-agents.
Instead, dependencies between sub-agents are implicit in their input and output artifact labels: an artifact produced by one sub-agent may satisfy an input required by another.
The Director is responsible for inferring these dependencies during planning.

At planning time, the descriptors of all registered sub-agents are rendered verbatim into the Director prompt as a capability catalog, preceded by the routing rules and the list of allowed identifiers.
The same descriptor representation is used during both training and inference.
In the configuration of Section~\ref{sec:experiments}, individual descriptors range from approximately 0.6K to 3.8K characters, while the full catalog contains about 30K characters.
Thus, the prompt is dominated by sub-agent descriptors rather than by handcrafted examples, and it contains no example plans.
\figurename~\ref{fig:descriptor_supp} shows one descriptor exactly as rendered into the Director prompt.
The descriptor schema and the sub-agent implementations will be released.

{
\captionsetup{type=figure}
\begin{fileviewerbox}{A sub-agent descriptor as rendered into the Director prompt}
\begin{Verbatim}[breaklines=true,fontsize=\scriptsize]
StoryAgent
  - Inputs:
      [creative_brief] (single, required): A natural-language brief describing what kind of story / video to produce. May be a short prompt ('a film about a cat chasing a butterfly') or a longer detailed outline / draft story text. The caption describes it as a creative brief / project intent description. If an enriched brief (one that integrates image reference descriptions) is available, prefer it over the raw text-intake brief. Pick the single most recent / most authoritative such brief.
      [reference_analysis] (single, optional): Optional scene-level video-analysis report of an inspiration / reference video. Use cases include 'analyse this hit drama and write me a same-genre sequel / similar new story' — the LLM reads its genre, mood, scene summaries, and entities as creative seeds to shape tone and pacing of the NEW blueprint, without copying plot verbatim. Skip this label when there is no reference video being analysed.
  - Output: story_blueprint (logline, cast, locations, story_arc, scene_outline).
  - Purpose / Trigger: First creative step for any task that produces a NEW film from scratch. Trigger: requests to create new film content (mini-drama, manhua, trailer, vertical short, etc.) — user provides a creative brief and is not asking to transform / edit an existing source video. A reference video used purely as creative inspiration (for genre / mood seeding via an upstream video analysis report) is allowed and does not disqualify this trigger.
\end{Verbatim}
\end{fileviewerbox}
\captionof{figure}{
A sub-agent descriptor exactly as rendered into the Director prompt (\texttt{Inputs}, \texttt{Output}, \texttt{Purpose / Trigger}).
}
\label{fig:descriptor_supp}
}

\subsection{Training and Reproducibility Details}
\label{sec:repro_supp}

\noindent\textbf{Director training.}
The Director is initialized from Qwen3-8B and fine-tuned with LoRA adapters (rank $32$, $\alpha=64$, dropout $0.05$; applied to all attention and MLP projections), with reasoning (``thinking'') disabled. For SFT we use a learning rate of $2\times10^{-5}$, an effective batch size of $8$ (per-device batch $1$, gradient accumulation $4$, $2$ GPUs), $4$ epochs, and a maximum sequence length of $12{,}288$ tokens; the loss is computed on assistant response tokens only, with the prompt masked out.
For GRPO we use group size $G=4$, reward weights $\lambda_{\mathrm{val}}=0.1$, $\lambda_{\mathrm{cov}}=0.5$, $\lambda_{\mathrm{ord}}=0.4$, KL coefficient $\beta=0.04$ (K3 estimator), sampling temperature $0.9$ (top-$p=1.0$, top-$k$ disabled), maximum $1{,}024$ new tokens, and a learning rate of $1\times10^{-6}$, optimizing against a frozen SFT reference policy for $4$ epochs. Updates are single-step and on-policy (one gradient step per sampled group).

\noindent\textbf{Backends and versions.}
Unless otherwise stated, the Assistant and sub-agents use Gemini~2.5 Flash as the internal reasoning backend; the end-to-end generation backends are \texttt{gemini-3-pro-image-preview} (2K resolution) for text-to-image and a self-hosted HunyuanVideo-I2V server ($25$ denoising steps) for image-to-video. CLIP scores use ViT-B/32, and the MLLM judge is Gemini~2.5 Pro.

\subsection{Prompt for the MLLM Judge}
\label{MLLM}
For the MLLM-as-a-judge, we use the following prompt:

{
\captionsetup{type=figure}
\begin{fileviewerbox}{Prompt for MLLM-as-a-judge}
\begin{Verbatim}[breaklines=true]

You are a rigorous multi-modal video evaluation expert (MLLM as a judge). Based only on the provided frames/timestamps and text/control information, evaluate a single video with structured scoring and traceable evidence. Do not hallucinate unseen content.

C1. Semantic Content Accuracy (Objects & Scene)
- What to check:
  - Are the specified object categories present and correct?
  - Is the overall scene type (nature/city/indoor/outdoor/weather/terrain) correct and stable?
  - Are major scene elements consistent with the text prompt and visual context?
- Typical evidence: timestamps where required objects/scenes appear (or fail), brief notes on correctness.
- Anchors:
1: Objects/scenes largely wrong or missing; persistent mismatch in most segments.
2: Frequent mismatches; objects or scene type often incorrect or unstable.
3: Mostly correct but with noticeable lapses (e.g., brief wrong class or scene drift).
4: Correct and stable with only minor slips in a few moments.
5: Fully correct and stable throughout; no contradictory frames observed.

C2. Multi-Object & Spatial Relations
- What to check:
  - Are object counts correct and stable?
  - Are spatial relations (above/below, inside/outside, left/right, front/back) correct and perspective-consistent?
  - Are occlusions and object interactions physically plausible?
- Typical evidence: frames showing relation satisfaction/violation (e.g., "cup above plate").
- Anchors:
1: Major errors in count/placement; relations frequently wrong or contradictory.
2: Multiple wrong relations or unstable layouts; occlusion frequently implausible.
3: Largely correct with occasional conflicts or transient misplacements.
4: Almost entirely correct; rare, minor inconsistencies.
5: Fully correct and stable; relations clear and consistently maintained.

C3. Action / Behavior Accuracy (Human or Specified Agent)
- What to check:
  - If the prompt specifies actions or poses (e.g., "running", "waving"), are they recognizable and temporally continuous?
  - Are motion trajectories and behavior transitions coherent?
  - Is the intended action maintained throughout the relevant duration?
  - If no action is specified, set null.
- Typical evidence: timestamps covering onset, continuity, and completion of the action.
- Anchors:
1: Action absent or clearly wrong most of the time.
2: Frequent mismatches or fragmentation; hard to recognize the intended action.
3: Generally matches, but with noticeable distortions or brief interruptions.
4: Clear and continuous match, with minor imperfections only.
5: Strong, consistent match; clear start-to-end execution with no ambiguity.

C4. Story Richness & Scene Diversity
- What to check:
  - Does the video present diverse scenes, locations, or environments rather than repetitive content?
  - Is there meaningful narrative or event progression across the video?
  - Are transitions between scenes, events, or cinematic moments coherent and visually engaging?
  - Does the video maintain viewer interest through evolving visual storytelling rather than static repetition?
- Typical evidence: timestamps showing scene transitions, new environments, evolving events, or repeated/redundant content.
- Anchors:
1: Extremely repetitive or static; little to no scene variation or narrative progression.
2: Limited diversity; scenes or events repeat frequently with weak progression.
3: Moderate scene diversity and storytelling progression, but with noticeable repetition or underdeveloped transitions.
4: Rich and coherent storytelling with diverse scenes and only minor repetition or transition issues.
5: Highly engaging and diverse visual storytelling with coherent progression, varied scenes, and strong cinematic development throughout.

C5. Style Consistency (Appearance & Cinematic Movement)
- What to check:
  - Does the visual appearance style (e.g., oil painting, cyberpunk, monochrome) match the prompt and remain temporally consistent?
  - Do camera movements (zoom, pan, dolly, tilt, handheld, etc.) match the intended cinematic style?
  - Are visual aesthetics and motion grammar stable throughout the video?
- Typical evidence: timestamps showing style adoption, camera behavior, or temporal drift.
- Anchors:
1: Style severely mismatched or mostly absent; camera grammar opposite or missing.
2: Frequent mismatches or drift in either appearance or camera style.
3: Generally matches with occasional drift or brief instability.
4: Clear and consistent match with only slight, rare issues.
5: Fully consistent in both appearance and camera grammar throughout.

C6. Overall Video-Text Consistency (set null if no text prompt)
- What to check:
  - Does the video holistically satisfy the text prompt?
  - Are scene, actions, style, objects, and narrative progression mutually coherent?
  - Does the overall semantic impression align with the intended theme and user instruction?
  - Avoid double-counting fine-grained issues already covered above.
- Typical evidence: timestamps representing core theme fulfillment or contradictions.
- Anchors:
1: Largely mismatched; core theme or requirements not met.
2: Many inconsistencies across key elements (theme/scene/action/style).
3: Mostly correct with noticeable errors in secondary aspects.
4: Overall consistent with small mismatches that do not change the theme.
5: Highly consistent; strong semantic agreement with the text prompt.



[Reference text for evaluation]
{target_text}

[Media context]
{media_description}

Respond with a JSON object containing at minimum a numeric ``score`` and a
string ``reasoning``. Optional ``subscores`` may carry per-dimension C1..C6
scores. Do not wrap the JSON in extra commentary.


\end{Verbatim}
\end{fileviewerbox}
\captionof{figure}{
Prompt for MLLM-as-a-judge.
}
\label{fig:mllm_prompt}
}

\section{User Study}
\label{sec:user-study}

\subsection{Rating Instrument}
\label{sec:rating-instrument}

Each \emph{case--system} tuple was rated on a 1--5 Likert scale along three dimensions, each decomposed into three sub-questions, nine in total. For each dimension, raters first saw a short hint and then answered each sub-question independently. All definitions were displayed in the rater's selected language. We show the English version below.

\paragraph{Instruction following.}
This dimension measures adherence to the input prompt. Controllability-related phrases in the prompt, including BGM, subtitles, ambient sound, style tags, and admin-curated keywords, were highlighted in red to anchor the rater's attention.

\begin{itemize}
    \item \textbf{Controllability.}
    Does the video respond to explicit control signals in the prompt, such as BGM, subtitles, ambient sound, and style tags? A score of 5 indicates that every requested control signal is clearly produced, whereas a score of 1 indicates that none of them appear.

    \item \textbf{Prompt adherence.}
    Does the video content match the prompt, including the described subjects, actions, setting, time, and story arc? When both a text prompt and a reference image were provided, raters were told to prioritize the text prompt over the reference image.

    \item \textbf{Absence of irrelevant frames.}
    Does the video remain on-prompt throughout its entire duration? Higher scores indicate fewer frames or sequences that were not requested by the prompt.
\end{itemize}

\paragraph{Temporal consistency.}
This dimension measures the stability of the depicted world across frames. Raters were asked to focus on how characters, objects, and the environment evolve over time.

\begin{itemize}
    \item \textbf{Subject stability.}
    Faces, identities, and body proportions of characters remain consistent across frames.

    \item \textbf{Object stability.}
    Props, clothing, and objects keep their shape, appearance, and position over time.

    \item \textbf{Environment stability.}
    The background, lighting, and scene layout remain coherent across frames, without brightness, color, or texture flicker.
\end{itemize}

\begin{figure}[h]
    \centering
    \includegraphics[width=\linewidth]{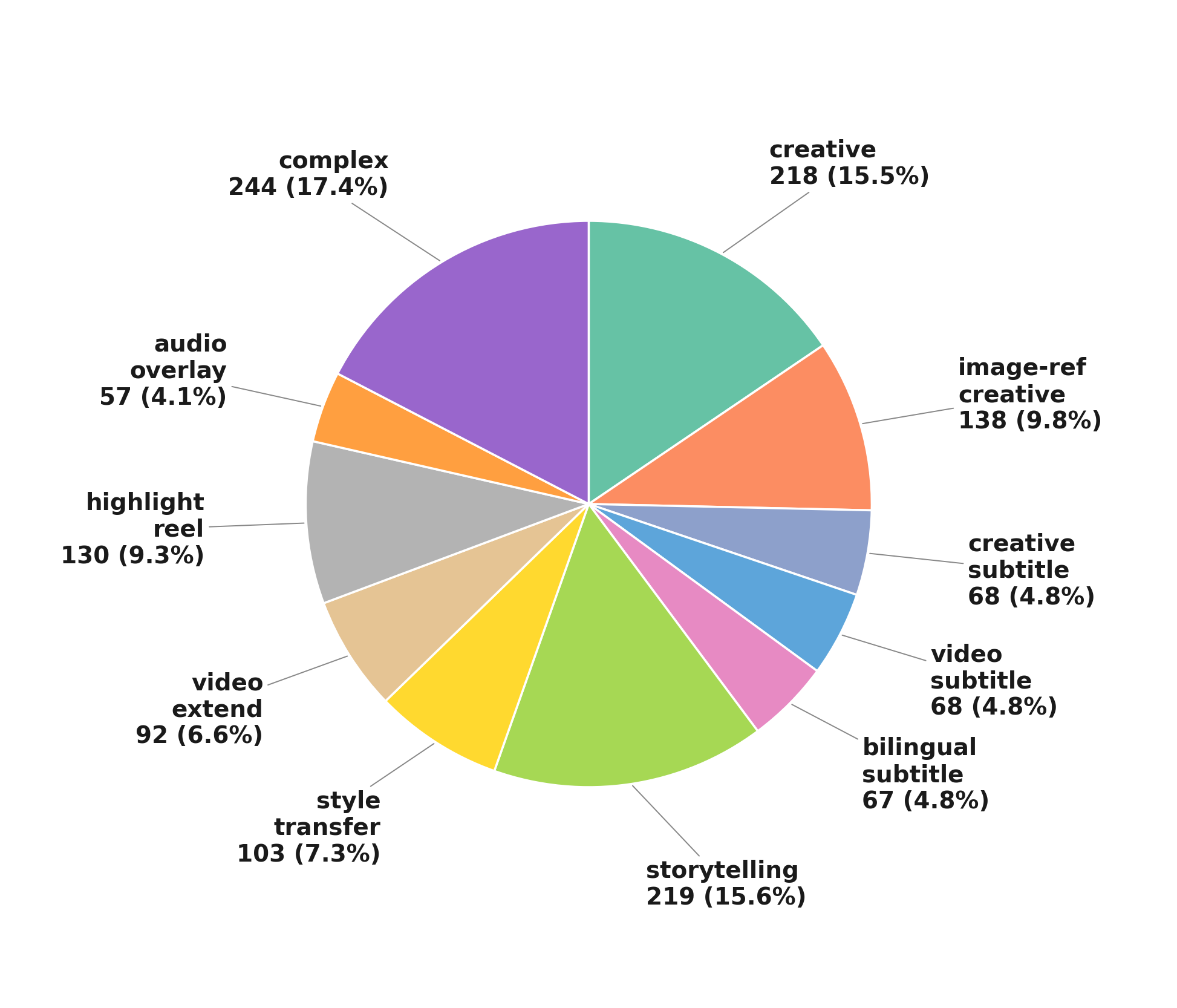}
    \caption{Category distribution of the routing training corpus.}
    \Description{}
    \label{fig:sft-category}
\end{figure}

\paragraph{Realism.}
This dimension measures how natural the video looks and moves. Raters were instructed to penalize obvious out-of-place or unnatural moments.

\begin{itemize}
    \item \textbf{Static appearance.}
    Per-frame photorealism or stylistic fidelity. When the prompt specifies a style, such as animated short drama or wuxia/xianxia (Chinese martial-arts / immortality-fantasy genres), the video is judged against that style rather than against photorealism. Uncanny or broken artifacts lower the score.

    \item \textbf{Motion plausibility.}
    Body and object motion should flow smoothly and follow physical and logical patterns, including gravity, collisions, and cloth or fluid dynamics. Warping, teleportation, and impossible movements lower the score. Videos with little or no motion receive a low score; this was made explicit in the rubric so that raters do not reward static outputs.

    \item \textbf{Camera motion.}
    Camera movement, such as pans, dollies, or focal changes, should feel natural and intentional. As with motion plausibility, little or no camera movement receives a low score; this was stated explicitly to avoid rewarding locked-off cameras.
\end{itemize}

\subsection{Operational Notes for Raters}
\label{sec:operational-notes}

Beyond the rubric, raters received three procedural reminders through an in-app help popup, which opened automatically once per device and could be reopened from a ``?'' button:

\begin{itemize}
    \item For instruction-following questions, read the prompt carefully, with special attention to the red-highlighted phrases, before scoring.
    \item For temporal-consistency questions, watch how persons, objects, and the environment change across the video.
    \item For realism questions, listen to the audio track when BGM, subtitles, or ambient sound are mentioned, and note any moment that feels obviously out of place.
\end{itemize}

Raters could save their progress at any time. Only sessions explicitly submitted by the rater were included in the analysis.


\end{document}